\documentclass[twoside]{article}

\usepackage[accepted]{aistats2018}
\usepackage{natbib}

\usepackage{hyperref}
\usepackage{booktabs}
\usepackage{multirow}
\usepackage{color}
\usepackage{ifthen}

\usepackage{subfigure} 
\usepackage{stackengine}

\usepackage{algorithm}
\usepackage{algpseudocode}

\usepackage{amssymb,amsmath}
\usepackage{mathbbol}
\usepackage{mathtools}  
\mathtoolsset{showonlyrefs}  
\usepackage{bm}
\usepackage{nicefrac}       

\newcommand{\tolowercase}[1]
{
{\ifthenelse{\equal{#1}{\Pi}}{\pi}{\MakeLowercase{#1}}}
}

\newcommand\bs[1]{\bm{#1}}
\newcommand\tr{\top}
\DeclareMathOperator*{\dom}{dom}

\newcommand\col[2]{\bs{\tolowercase{#1}}_{#2}}
\newcommand\matelem[3]{\tolowercase{#1}_{#2,#3}}
\DeclareMathOperator*{\argmin}{argmin}
\DeclareMathOperator*{\argmax}{argmax}
\newcommand{\Simplex}{\triangle}

\newcommand\va{\bs{a}}
\newcommand\vb{\bs{b}}
\newcommand\vx{\bs{x}}
\newcommand\vy{\bs{y}}
\newcommand\valpha{\bs{\alpha}}
\newcommand\vbeta{\bs{\beta}}
\newcommand\U{\mathcal{U}}
\newcommand\T{T}
\newcommand\KL{\text{KL}}
\newcommand\OT{\text{OT}}
\newcommand\OTomega{\text{OT}_{\Omega}}
\newcommand\OTphi{\text{ROT}_{\Phi}}
\newcommand\OTphir{\widetilde{\text{ROT}}_{\Phi}}
\newcommand\wrt{w.r.t.\ }
\newcommand\indneg{\delta}
\newcommand\indomega{\delta_{\Omega}}
\newcommand\maxop{\text{max}_{\Omega}}
\newcommand\maxopj{\text{max}_{\Omega_j}}
\newcommand\expdivgamma[2]{e^{\frac{#1}{\gamma} #2}}

\newcommand{\normtwo}[1]{|\!| #1 |\!|_2}
\newcommand{\normone}[1]{|\!| #1 |\!|_1}
\newcommand{\norminf}[1]{\left|\!\left| #1 \right|\!\right|_\infty}

\newtheorem{theorem}{Theorem} 
\newtheorem{lemma}{Lemma} 
\newtheorem{proposition}{Proposition}

\newtheorem{definition}{Definition}

\begin{document}


\runningauthor{Mathieu Blondel, Vivien Seguy, Antoine Rolet}

\twocolumn[

\aistatstitle{Smooth and Sparse Optimal Transport}

\aistatsauthor{Mathieu Blondel \And Vivien Seguy\textsuperscript{*} \And Antoine
Rolet\textsuperscript{*}}
\aistatsaddress{NTT Communication Science Laboratories \And Kyoto University \And  Kyoto University} 


]

\begin{abstract}
Entropic regularization is quickly emerging as a new standard in optimal
transport (OT). It enables to cast the OT computation as a
\textit{differentiable} and
\textit{unconstrained} convex optimization problem, which can be efficiently
solved using the Sinkhorn algorithm. However, entropy keeps the
transportation plan strictly positive and therefore completely dense, unlike
unregularized OT. This lack of sparsity can be problematic in applications
where the transportation plan itself is of interest. In this paper, we explore
regularizing the primal and dual OT formulations with a 
\textit{strongly convex} term, which corresponds to relaxing the dual and primal
constraints with \textit{smooth} approximations. We show how
to incorporate squared $2$-norm and group lasso regularizations within that
framework, leading to \textit{sparse} and \textit{group-sparse} transportation
plans.  On the theoretical side, we bound the approximation error
introduced by regularizing the primal and dual formulations. Our results
suggest that, for the regularized primal, the approximation error can often be
smaller with squared $2$-norm than with entropic regularization.
We showcase our proposed framework on the task of color transfer.
\end{abstract}

\begin{figure*}[t]
\centering
\scriptsize
\stackunder[2pt]{
\stackunder[5pt]{
    \includegraphics[scale=0.43]{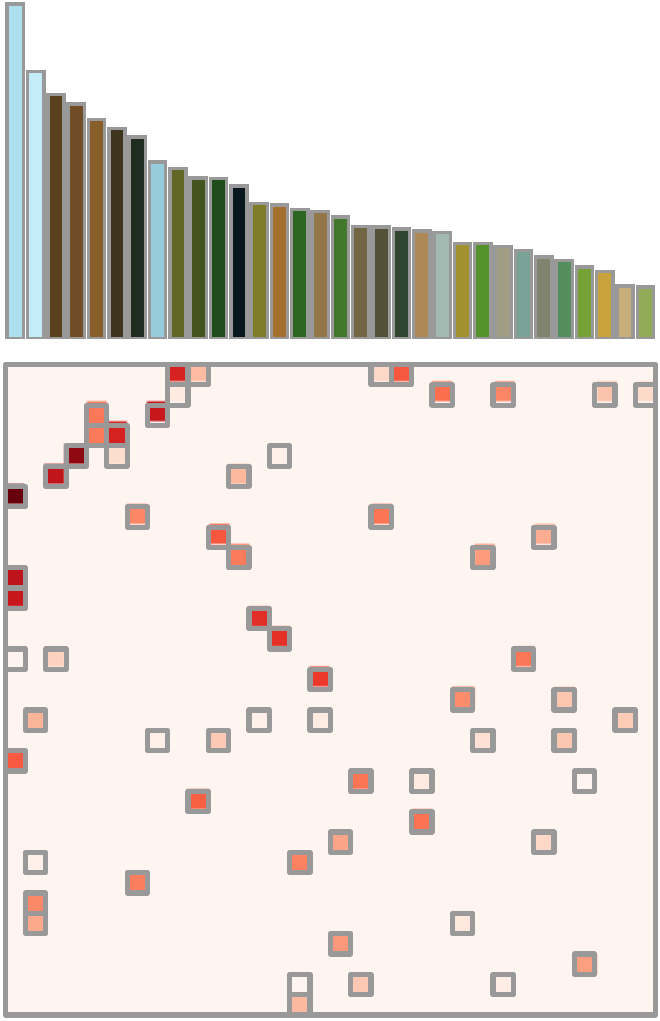}
}{Unregularized}}{Sparsity: 94\%}
\hspace{0.1cm}
\stackunder[2pt]{
\stackunder[5pt]{
    \includegraphics[scale=0.43]{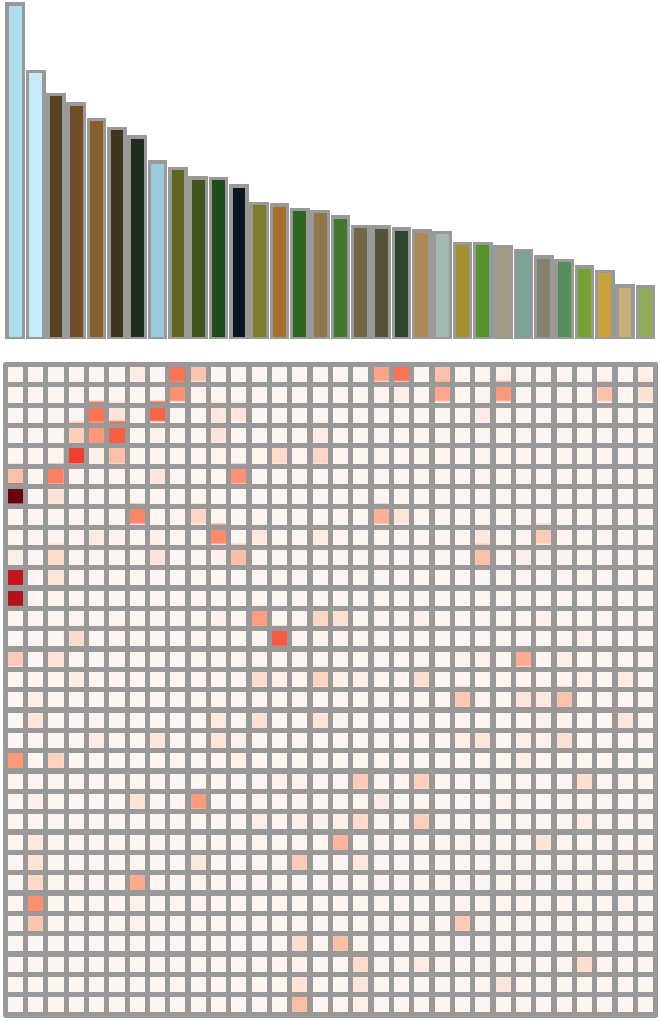}
}{Smoothed semi-dual (ent.)}}{Sparsity: 0\%}
\hspace{0.1cm}
\stackunder[2pt]{
\stackunder[5pt]{
    \includegraphics[scale=0.43]{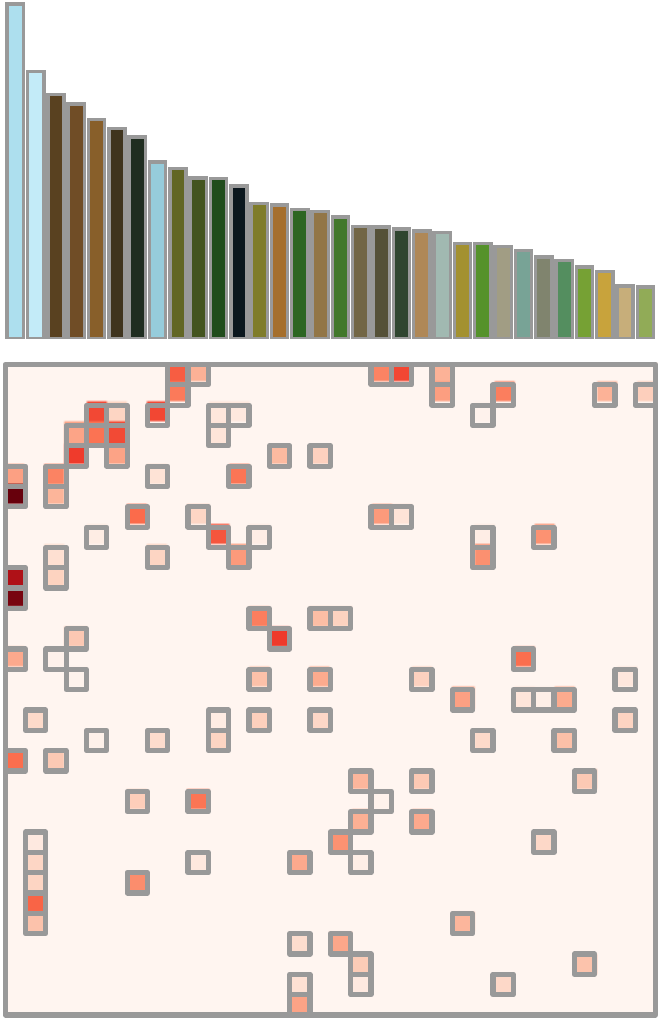}
}{Smoothed semi-dual (sq.\ $2$-norm)}}{Sparsity: 90\%}
\hspace{0.1cm}
\stackunder[2pt]{
\stackunder[5pt]{
\includegraphics[scale=0.43]{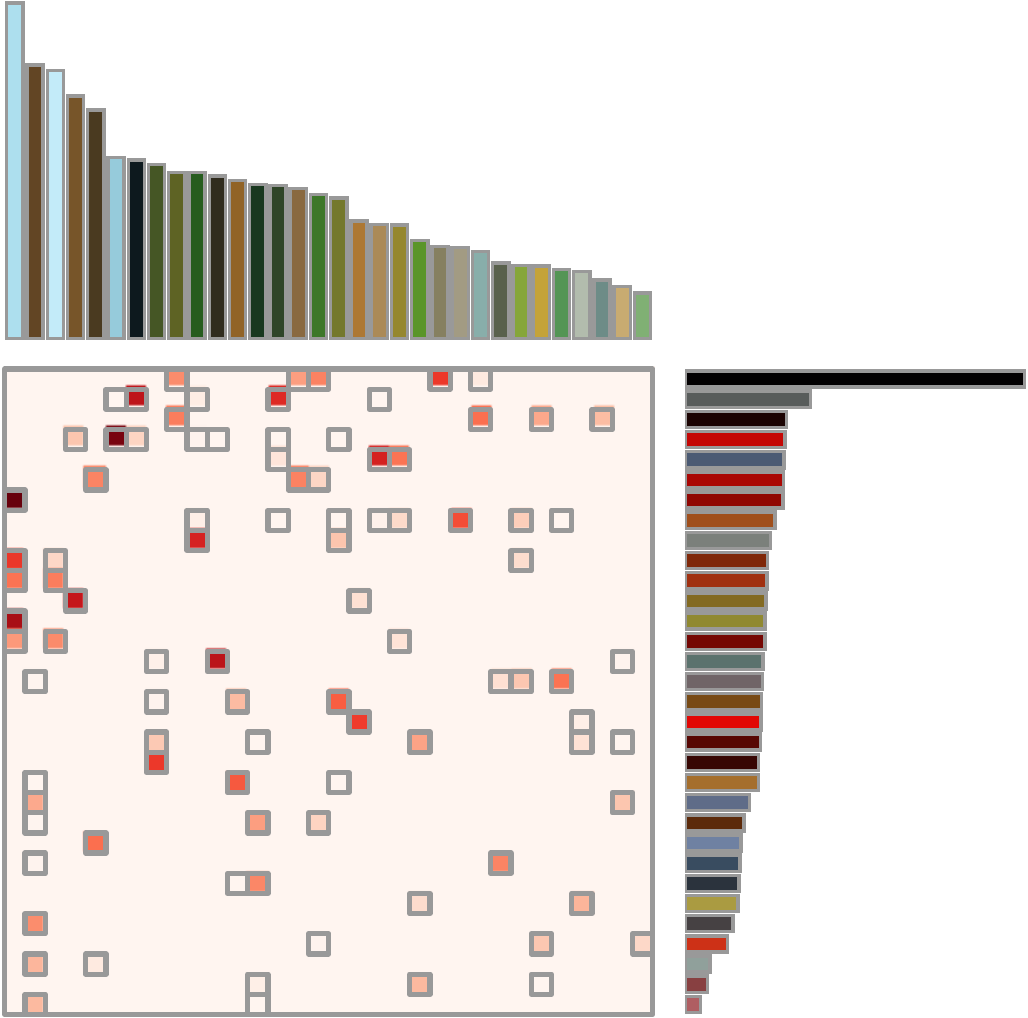}
}{Semi-relaxed primal (Eucl.)}}{Sparsity: 91\%}
\caption{Comparison of transportation plans obtained by different formulations on
    the application of color transfer. The top and right histograms represent
    the color distributions $\bs{a} \in \Simplex^m$ and $\bs{b} \in \Simplex^n$ of
    two images. For the sake of illustration, the
    number of colors is reduced to $m=n=32$, using k-means clustering. Small squares
    indicate non-zero elements in the obtained transportation plan, denoted by
    $\T$ throughout this paper. The sparsity indicated below each graph is the
    percentage of zero elements in $\T$. 
    The weight of the elements of $\T$ indicates the extent to which colors
    from one image must be transferred to colors from the other image. Like
    unregularized OT (first from left), but unlike entropy-regularized OT
    (second from left), our squared $2$-norm
    regularized OT (third from left) is able to produce sparse transportation plans. This
    is also the case of our relaxed primal (not shown) and semi-relaxed primal
    (fourth from left) formulations.
}
\label{fig:transp_plans}
\end{figure*}

\section{Introduction}

Optimal transport (OT) distances (a.k.a. Wasserstein or earth mover's distances)
are a powerful computational tool to compare probability distributions and have
recently found widespread use in machine learning
\citep{sinkhorn_distances,w_propagation,word_mover,ot_domain_adapt,wgan}.  While
OT distances exhibit a unique ability to capture the geometry of the data, their
application to large-scale problems has been largely hampered by their high
computational cost. Indeed, computing OT distances involves a linear program, which
takes super-cubic time in the data size to solve using state-of-the-art
network-flow algorithms.  Related to the Schr\"{o}dinger problem
\citep{schrodinger,survey_schrodinger}, entropy-regularized OT distances have
recently gained popularity due to their desirable properties
\citep{sinkhorn_distances}. Their computation
involves a comparatively easier \textit{differentiable} and
\textit{unconstrained} convex optimization problem, which can be solved using
the Sinkhorn algorithm \citep{sinkhorn}.  Unlike
unregularized OT distances, entropy-regularized OT distances are also
differentiable \wrt their inputs, enabling their use as a loss function in a
machine learning pipeline \citep{w_loss,paper_antoine}.

Despite its considerable merits, however, entropy-regularized OT has some limitations,
such as introducing blurring in the optimal transportation plan. While
this nuisance can be reduced by using small regularization, this requires a
carefully engineered implementation, since the naive Sinkhorn algorithm is
numerically unstable in this regime \citep{stabilized_sinkhorn}. More
importantly, the entropy term keeps the transportation plan strictly positive
and therefore completely dense, unlike unregularized OT. This lack of sparsity
can be problematic when the optimal transportation plan itself is of interest,
\textit{e.g.}, in color transfer \citep{color_transfer}, domain adaptation
\citep{ot_domain_adapt} and ecological inference \citep{tsallis_ot}.
Sparsity in these applications is motivated by the principle of parsimony
(simple solutions should be preferred) and by the enhanced interpretability of
transportation plans.

\textbf{Our contributions.} This background motivates us to study regularization
schemes that lead to \textit{smooth}  optimization problems (\textit{i.e.},
differentiable everywhere and with Lipschitz continuous gradient) while
retaining the desirable property of \textit{sparse} transportation plans. To do
so, we make the following contributions.

We regularize the primal with an arbitrary strongly convex term and
derive the corresponding smoothed dual and semi-dual.  Our derivations abstract
away regularization-specific terms in an intuitive way
(\S\ref{sec:smoothed_dual}).  We show how incorporating squared $2$-norm and
group-lasso regularizations within that framework leads to sparse solutions.
This is illustrated in Figure
\ref{fig:transp_plans} for squared $2$-norm regularization.

Next, we explore the opposite direction: replacing one or both of the
primal marginal constraints with approximate smooth constraints.  
When using the squared
Euclidean distance to approximate the constraints, we show that this can be
interpreted as adding squared
$2$-norm regularization to the dual (\S\ref{sec:relaxed_primal}).
As illustrated in Figure \ref{fig:transp_plans}, that approach also produces
sparse transportation plans.  

For both directions, we bound the approximation error caused by regularizing the
original OT problem.  For the regularized primal, we show that the approximation
error of squared $2$-norm regularization can be smaller than that of entropic
regularization (\S\ref{sec:bounds}).
Finally, we showcase the proposed approaches empirically on the task of color
transfer (\S\ref{sec:exp}).

An open-source Python implementation is available at
\url{https://github.com/mblondel/smooth-ot}.

\textbf{Notation.} We denote scalars, vectors and matrices using lower-case,
bold lower-case and upper-case letters, \textit{e.g.}, $t$, $\bs{t}$ and $\T$,
respectively.  
Given a matrix $\T$, we denote its elements by $\matelem{T}{i}{j}$ and its columns
by $\col{T}{j}$.
We denote the set $\{1,\dots,m\}$ by $[m]$.  
We use $\|\cdot\|_p$ to denote the $p$-norm. When $p=2$, we simply write
$\|\cdot\|$.
We denote the $(m-1)$-dimensional probability
simplex by $\Simplex^m \coloneqq \{ \bs{y} \in \mathbb{R}^m_+ \colon \|\bs{y}\|_1 =
1\}$ and the Euclidean projection onto it by
$P_{\Simplex^m}(\bs{x}) \coloneqq \argmin_{\bs{y} \in \Simplex^m} \|\bs{y} -
\bs{x}\|^2$. We denote $[\bs{x}]_+ \coloneqq \max(\bs{x}, \bs{0})$, performed
element-wise.

\section{Background}

\textbf{Convex analysis.}
The convex conjugate of a function $f \colon \mathbb{R}^m \to \mathbb{R} \cup
\{\infty\}$ is defined by
\begin{equation}
f^*(\bs{x}) \coloneqq \sup_{\bs{y} \in \dom f} ~ \bs{y}^\tr \bs{x} - f(\bs{y}).
\label{eq:conjugate}
\end{equation}
If $f$ is strictly convex, then the
supremum in \eqref{eq:conjugate} is uniquely achieved.
Then, from Danskin's theorem (\citeyear{danskin_theorem}),
it is equal to the gradient of $f^*$:
\begin{equation}
\nabla f^*(\bs{x}) = \argmax_{\bs{y} \in \dom f} ~ \bs{y}^\tr \bs{x} - f(\bs{y}).
\label{eq:grad_conjugate}
\end{equation}
The dual of a norm $\|\cdot\|$ is defined by
$
\|\bs{x}\|_* \coloneqq \sup_{\|\bs{y}\| \le 1} ~ \bs{y}^\tr \bs{x}.
$
We say that a function is $\gamma$-smooth \wrt a norm $\|\cdot\|$ if it
is differentiable everywhere and its gradient is $\gamma$-Lipschitz continuous
\wrt that norm.  Strong convexity plays a crucial role in this paper due to
its well-known duality with smoothness: $f$ is
$\gamma$-strongly convex \wrt a norm $\|\cdot\|$ if and only if $f^*$ is
$\frac{1}{\gamma}$-smooth \wrt $\|\cdot\|_*$ 
\citep{kakade_jmlr}.  

\textbf{Optimal transport.} We focus throughout this paper on OT between
discrete probability distributions $\bs{a} \in \Simplex^m$ and $\bs{b} \in
\Simplex^n$. 
Rather than performing a pointwise comparison
of the distributions, OT distances compute the minimal effort, according to some
ground cost, for moving the probability mass of one distribution to the other.
The modern OT formulation, due to Kantorovich [\citeyear{kantorovich}], is cast
as a linear program (LP):
\begin{equation}
\text{OT}(\bs{a}, \bs{b}) \coloneqq
\min_{\T \in \mathcal{U}(\bs{a},\bs{b})} \langle \T, C \rangle,
\label{eq:primal_lp}
\end{equation}
where $\mathcal{U}(\bs{a}, \bs{b})$ is the transportation polytope
\begin{equation}
    \mathcal{U}(\bs{a},\bs{b}) \coloneqq \left\{ \T \in \mathbb{R}^{m \times n}_+ \colon
\T \bs{1}_n = \bs{a}, \T^\tr \bs{1}_m = \bs{b} \right\}
\end{equation}
and $C \in \mathbb{R}_+^{m \times n}$ is a cost matrix. The former can be
interpreted as the set of all joint probability distributions with marginals
$\bs{a}$ and $\bs{b}$. 
Without loss of generality, we will assume $\bs{a} > 0$ and $\bs{b}
> 0$ throughout this paper (if $a_i=0$ or $b_j=0$ then the
$i$\textsuperscript{th} row or the $j$\textsuperscript{th} column of $T^\star$
is zero).
When $n=m$ and $C$ is a distance matrix raised to the power $p$, $\OT(\cdot,
\cdot)^{\frac{1}{p}}$ is a distance on $\Simplex^n$, called the Wasserstein
distance of order $p$ \citep[Theorem 7.3]{villani_2003}.
The dual LP is
\begin{equation}
\text{OT}(\bs{a}, \bs{b}) =
\max_{\bs{\alpha}, \bs{\beta} \in \mathcal{P}(C)}
\bs{\alpha}^\tr \bs{a} + \bs{\beta}^\tr \bs{b},
\label{eq:dual_lp}
\end{equation}
where 
$
    \mathcal{P}(C) \coloneqq \{\bs{\alpha} \in \mathbb{R}^m, \bs{\beta} \in \mathbb{R}^n
\colon \alpha_i + \beta_j \le \matelem{C}{i}{j}\}.
$
Keeping $\bs{\alpha}$ fixed, an optimal solution \wrt $\bs{\beta}$ is
\begin{equation}
    \beta_j = \min_{i \in [m]} \matelem{C}{i}{j} - \alpha_i,
\quad \forall j \in [n],
\end{equation}
which is the so-called \textit{c-transform}.
Plugging it back into the dual, we get the ``semi-dual''
\begin{equation}
\text{OT}(\bs{a}, \bs{b}) =
\max_{\bs{\alpha} \in \mathbb{R}^m} 
\bs{\alpha}^\tr \bs{a} - \sum_{j=1}^n b_j 
\max_{i \in [m]}(\alpha_i - \matelem{C}{i}{j}).
\label{eq:semi_dual}
\end{equation}
For a recent and comprehensive survey of computational OT, see
\citep{peyre2017computational}.

\section{Strong primal $\leftrightarrow$ Relaxed dual}
\label{sec:smoothed_dual}

We study in this section adding strongly convex regularization to the primal
problem \eqref{eq:primal_lp}. We define
\begin{definition}{Strongly convex primal}
\begin{equation}
\normalfont
\OTomega(\bs{a}, \bs{b}) \coloneqq 
\min_{\T \in \mathcal{U}(\bs{a}, \bs{b})} 
\langle \T, C \rangle 
+ \sum_{j=1}^n \Omega(\col{\T}{j}),
\label{eq:regularized_primal}
\end{equation}
where we assume that $\Omega$ is strongly convex over the intersection of $\dom
\Omega$ and either $\mathbb{R}^m_{+}$ or $\Simplex^m$.
\end{definition}
These assumptions are sufficient for \eqref{eq:regularized_primal} to be
strongly convex \wrt $\T \in \U(\va,\vb)$.  On first sight, solving
\eqref{eq:regularized_primal} does not seem easier than \eqref{eq:primal_lp}.
As we shall now see, the main benefit occurs when switching to the dual.

\subsection{Smooth relaxed dual formulation}

Let the (non-smooth)
indicator function of the non-positive orthant be defined as
\begin{equation}
\indneg(\bs{x}) \coloneqq
\begin{cases} 
0, & \mbox{if }  \bs{x} \le 0  \\ 
\infty,  & \mbox{o.w. } 
\end{cases} 
= \sup_{\bs{y} \ge 0} \bs{y}^\tr \bs{x}.
\end{equation}
To define a smoothed version of $\delta$, we take the convex conjugate of
$\Omega$, restricted to the non-negative orthant:
\begin{equation}
    \indomega(\bs{x}) \coloneqq 
\sup_{\bs{y} \ge 0} \bs{y}^\tr \bs{x} - \Omega(\bs{y}).
\label{eq:smoothed_indicator}
\end{equation}
If $\Omega$ is $\gamma$-strongly convex over $\mathbb{R}^m_{+} \cap \dom
\Omega$, then $\indomega$ is $\frac{1}{\gamma}$-smooth and its gradient is
$\nabla \indomega(\bs{x}) = \bs{y}^\star$, where $\vy^\star$ is the supremum of
\eqref{eq:smoothed_indicator}. 
We next show that $\indomega$ plays a
crucial
role in expressing the dual of \eqref{eq:regularized_primal}, which is a smooth
optimization problem in $\valpha$ and $\vbeta$. 
\begin{proposition}{Smooth relaxed dual}
\begin{equation}
\normalfont
\OTomega(\bs{a}, \bs{b})
= \max_{\substack{\bs{\alpha} \in \mathbb{R}^m\\ \bs{\beta} \in \mathbb{R}^n}}
\bs{\alpha}^\tr \bs{a} + \bs{\beta}^\tr \bs{b} 
-\sum_{j=1}^n \indomega(\bs{\alpha} + \beta_j \bs{1}_m - \col{c}{j})
\label{eq:smoothed_dual}
\end{equation}
The optimal solution $T^\star$ of \eqref{eq:regularized_primal} can be recovered
from $(\bs{\alpha}^\star, \bs{\beta}^\star)$ by
$
\col{\T}{j}^\star = 
\nabla \indomega(\bs{\alpha}^\star + \beta_j^\star \bs{1}_m - \col{c}{j})
\quad \forall j \in [n].
$
\end{proposition}
For a proof, see Appendix \ref{appendix:proof_dual}. Intuitively, the hard dual
constraints $\alpha_i + \beta_j - \matelem{c}{i}{j} \le 0 ~\forall i \in [m]
~\forall j
\in [n]$, which we can write
$\sum_{j=1}^n \delta(\bs{\alpha} + \beta_j \bs{1}_m - \col{c}{j})$, are now
relaxed with \textit{soft} ones by substituting $\delta$
with $\indomega$.

\begin{table*}[t]
\caption{Closed forms for $\indomega$ (used in smoothed dual),
$\maxopj$ (used in smoothed semi-dual) and their gradients.}
\label{table:expressions}
\begin{small}
\begin{center}
\begin{tabular}{c c c c c c}
\toprule
& $\Omega(\bs{y})$
& $\indomega(\bs{x})$ & $\nabla \indomega(\bs{x})$
& $\maxopj(\bs{x})$ & $\nabla \maxopj(\bs{x})$ \\
\midrule
\addlinespace[0.2em]
Negative entropy & 
$\gamma \displaystyle{\sum_{i=1}^m} y_i \log y_i$ &
$\gamma \displaystyle{\sum_{i=1}^m} \expdivgamma{x_i}{- 1}$ &
$\expdivgamma{\bs{x}}{-\bs{1}_m}$ &
$\gamma \log \displaystyle{\sum_{i=1}^m \expdivgamma{x_i}{}} 
- \gamma \log b_j$ &
$\frac{\expdivgamma{\bs{x}}{}}{\sum_{i=1}^m \expdivgamma{x_i}{}}$ \\
\addlinespace[0.2em]
Squared $2$-norm & 
$\frac{\gamma}{2} \| \bs{y} \|^2$ & 
$\frac{1}{2\gamma} \displaystyle{\sum_{i=1}^m} [x_i]_+^2$ & 
$\frac{1}{\gamma} [\bs{x}]_+$ & 
$\bs{x}^\tr \bs{y}^\star - \frac{\gamma b_j}{2} \|\bs{y}^\star\|^2$ & 
$\bs{y}^\star = P_{\Simplex^m}\left(\frac{\bs{x}}{\gamma b_j}\right)$ \\
\addlinespace[0.2em]
Group-lasso &
$\frac{\gamma}{2} \|\bs{y}\|^2 + \gamma \mu
\displaystyle{\sum_{G \in \mathcal{G}}} \|\bs{y}_G\|$ &
\eqref{eq:smoothed_indicator_gl} &
\eqref{eq:smoothed_indicator_gl_sol} &
\multicolumn{2}{c}{No closed form available} \\
\bottomrule
\end{tabular}
\end{center}
\end{small}
\end{table*}

\subsection{Smoothed semi-dual formulation}

We now derive the semi-dual of \eqref{eq:regularized_primal}, \textit{i.e.}, the dual
\eqref{eq:smoothed_dual} with one of the two variables eliminated.  
Without loss of generality, we proceed to eliminate $\vbeta$.
To do so, we use the notion of smoothed max operator. Notice that 
\begin{equation}
\max(\bs{x}) \coloneqq \max_{i \in [m]} x_i 
= \sup_{\bs{y} \in \Simplex^m} \bs{y}^\tr \bs{x}
\quad \forall \bs{x} \in \mathbb{R}^m.
\end{equation}
This is indeed true, since the supremum is always achieved at one of the simplex
vertices.  To define a smoothed max operator \citep{nesterov_smooth}, we
take the conjugate of $\Omega$, this time restricted to the simplex:
\begin{equation}
\maxop(\bs{x}) \coloneqq \sup_{\bs{y} \in \Simplex^m} \bs{y}^\tr \bs{x} -
\Omega(\bs{y}).
\label{eq:maxop}
\end{equation}
If $\Omega$ is $\gamma$-strongly convex over $\Simplex^m \cap \dom \Omega$, 
then $\maxop$ is $\frac{1}{\gamma}$-smooth and its gradient is defined by $\nabla
\maxop(\bs{x}) = \bs{y}^\star$, where $\vy^\star$ is the supremum of
\eqref{eq:maxop}.  We next show that $\maxop$ plays a crucial role
in expressing the conjugate of $\OTomega$.
\begin{lemma}{Conjugate of $\OTomega$ \wrt its first argument}
\begin{equation}
\normalfont
\small
\OTomega^*(\valpha, \bs{b}) = 
\sup_{\bs{a} \in \Simplex^m} \valpha^\tr \bs{a} - 
\OTomega(\bs{a}, \bs{b}) =
\sum_{j=1}^n b_j \maxopj(\valpha - \col{c}{j}),
\label{eq:ot_omega_conjugate}
\end{equation}
where 
\normalfont{$\Omega_j(\bs{y}) \coloneqq \frac{1}{b_j} \Omega(b_j \bs{y})$}.
\label{lemma:ot_omega_conjugate}
\end{lemma}
A proof is given in Appendix \ref{appendix:proof_conjugate}.
With the conjugate, we can now easily express the semi-dual of
\eqref{eq:regularized_primal},
which involves a smooth optimization problem in $\valpha$.
\begin{proposition}{Smoothed semi-dual}
\begin{equation}
\normalfont
\OTomega(\bs{a}, \bs{b}) = 
\max_{\bs{\alpha} \in \mathbb{R}^m} \bs{\alpha}^\tr \bs{a} - 
\OTomega^*(\bs{\alpha}, \bs{b}) 
\label{eq:smoothed_semi_dual}
\end{equation}
The optimal solution $T^\star$ of \eqref{eq:regularized_primal} can be recovered
from $\bs{\alpha}^\star$ by
{\normalfont
$
\col{\T}{j}^\star = b_j \nabla \maxopj(\bs{\alpha}^\star - \col{c}{j})
\quad \forall j \in [n].
$
}
\end{proposition}
\textit{Proof.} $\OTomega(\va, \vb)$ is a closed
and convex function of $\va$. Therefore,
$\OTomega(\va, \vb) = \OTomega^{**}(\va, \vb)$. $\square$

We can interpret this semi-dual as a variant of \eqref{eq:semi_dual}, where the
max operator has been replaced with its smoothed counterpart, $\maxopj$.  Note
that $\bs{\alpha}^\star$, as obtained by solving the smoothed dual
\eqref{eq:smoothed_dual} or semi-dual \eqref{eq:smoothed_semi_dual},
is the gradient of $\OTomega(\bs{a}, \bs{b})$ \wrt $\bs{a}$ when
$\bs{\alpha}^\star$ is unique or a sub-gradient otherwise.  This is useful when
learning with $\OTomega$ as a loss, as done with entropic regularization in
\citep{w_loss}.

\textbf{Solving the optimization problems.} The dual and semi-dual we derived
are unconstrained, differentiable and concave optimization problems. They can
therefore be solved using gradient-based algorithms, as long as we know
how to compute $\nabla \indomega$ and $\nabla \maxop$. In our experiments, we
use L-BFGS \citep{lbfgs}, for both the dual and semi-dual formulations.

\subsection{Closed-form expressions}
\label{sec:closed_form_expr}

We derive in this section closed-form expressions for $\indomega$, $\maxop$ and
their gradients
for specific choices of $\Omega$.

\textbf{Negative entropy.} We choose $\Omega(\bs{y}) = -\gamma H(\bs{y})$, where
$H(\bs{y}) \coloneqq -\sum_i y_i \log y_i$ is the entropy.  For that
choice, we get analytical expressions for $\indomega$, $\maxop$ and their
gradients (cf.  Table \ref{table:expressions}).  
Since $\Omega$ is $\gamma$-strongly
convex \wrt the $1$-norm over $\Simplex^m$ \citep[Lemma
16]{shalev_shwartz_thesis}, $\maxop$ is $\frac{1}{\gamma}$-strongly smooth \wrt
the $\infty$-norm. However, since $\Omega$ is only strictly convex
over $\mathbb{R}_{>0}^{m}$, $\indomega$ is differentiable but \textit{not} smooth.
The dual and semi-dual with this $\Omega$ were derived in 
\citep{marco_barycenter} and
\citep{ot_sag}, respectively.

Next, we present two choices of $\Omega$ that induce sparsity in transportation
plans. The resulting dual and semi-dual expressions are new, to our knowledge.

\textbf{Squared $2$-norm.} 
We choose $\Omega(\bs{y}) = \frac{\gamma}{2} \|\bs{y}\|^2$. 
We again obtain
closed-form expressions for $\indomega$, $\maxop$ and their gradients (cf. Table
\ref{table:expressions}). 
Since $\Omega$ is $\gamma$-strongly convex \wrt the $2$-norm
over $\mathbb{R}^m$, both $\indomega$ and $\maxop$ are
$\frac{1}{\gamma}$-strongly smooth \wrt the $2$-norm.  
Projecting a vector onto the simplex, as required to
compute $\maxop$ and its gradient, can be done exactly in worst-case $O(m \log m)$ time
using the algorithm of \citep{michelot} and in expected $O(m)$ time using
the randomized pivot algorithm of \citep{duchi}.  Squared $2$-norm
regularization can output exactly sparse transportation plans (the primal-dual
relationship for \eqref{eq:smoothed_dual} is $\matelem{\T}{i}{j}^\star
= \frac{1}{\gamma} [\alpha_i^\star + \beta_j^\star - c_{i,j}]_+$) and is
numerically stable without any particular implementation trick.

\textbf{Group lasso.} \citet{ot_domain_adapt} recently proposed to use
$\Omega(\bs{y}) = \gamma (\sum_i y_i \log y_i + \mu \sum_{G \in \mathcal{G}}
\|\bs{y}_G\|)$, where $\bs{y}_G$ denotes the subvector of $\bs{y}$ restricted to
the set $G$, and showed that this regularization improves accuracy in the
context of domain adaptation. 
Since $\Omega$ includes a negative entropy term,
the same remarks as for negative entropy apply
regarding the differentiability of $\indomega$ and smoothness of $\maxop$.
Unfortunately, a closed-form solution is available for neither
\eqref{eq:smoothed_indicator} nor \eqref{eq:maxop}.
However, since the log keeps $\bs{y}$ in the strictly positive orthant and 
$\|\bs{y}_G\|$ is differentiable everywhere in that orthant, we can use any
proximal gradient algorithm to solve these problems to arbitrary precision.

A drawback of this choice of $\Omega$, however, is that group sparsity is never
truly achieved. To address this issue, we propose to use $\Omega(\bs{y}) =
\gamma (\frac{1}{2} \|\bs{y}\|^2 + \mu \sum_{G \in \mathcal{G}}
\|\bs{y}_G\|)$ instead. For that choice, $\indomega$ is smooth and is equal to
\begin{equation}
    \indomega(\bs{x}) = \bs{x}^\tr \bs{y}^\star - \Omega(\bs{y}^\star),
\label{eq:smoothed_indicator_gl}
\end{equation}
where $\bs{y}^\star$ decomposes over groups $G \in \mathcal{G}$ and equals
\begin{equation}
\bs{y}^\star_G =
\argmin_{\textcolor{blue}{\bs{y}_G \ge 0}} \frac{1}{2} \|\bs{y}_G - \bs{x}_G / \gamma\|^2
+ \mu \|\bs{y}_G\| = \nabla \indomega(\bs{x})_G.
\label{eq:smoothed_indicator_gl_sol}
\end{equation}
As noted in the context of group-sparse NMF
\citep{group_sparse_nmf}, \eqref{eq:smoothed_indicator_gl_sol} admits a
closed-form solution
\begin{equation}
\bs{y}^\star_G 
= \argmin_{\textcolor{blue}{\bs{y}_G}} \frac{1}{2} \left\|\bs{y}_G - \bs{x}_G^+\right\|^2
+ \mu \|\bs{y}_G\| 
= \bigg[1 - \frac{\mu}{\|\bs{x}_G^+\|}\bigg]_+ \bs{x}_G^+,
\end{equation}
where we defined $\bs{x}^+ \coloneqq \frac{1}{\gamma} [\bs{x}]_+$.  We have thus
obtained an efficient way to compute exact gradients of $\indomega$, making it
possible to solve the dual using gradient-based algorithms.  In contrast,
\citet{ot_domain_adapt} use a generalized conditional gradient algorithm whose
iterations require expensive calls to Sinkhorn.  Finally, because
$\col{\T}{j}^\star = \nabla \indomega(\bs{\alpha}^\star + \beta_j^\star
\bs{1}_m - \col{c}{j}) ~ \forall j \in [n]$, the obtained transportation plan
will be truly group-sparse. 

\section{Relaxed primal $\leftrightarrow$ Strong dual}
\label{sec:relaxed_primal}

We now explore the opposite way to define smooth OT problems while retaining
sparse transportation plans: replace marginal constraints in the primal with
approximate constraints. When relaxing both marginal constraints, we
define the next formulation:
\begin{definition}{Relaxed smooth primal}
\begin{equation}
\normalfont
\OTphi(\bs{a}, \bs{b}) \coloneqq 
\min_{\T \ge 0} ~
\langle \T, C \rangle 
+ \frac{1}{2} \Phi(\T \bs{1}_n, \bs{a})
+ \frac{1}{2} \Phi(\T^\tr \bs{1}_m, \bs{b}),
\label{eq:relaxed_primal}
\end{equation}
where $\Phi(\bs{x}, \bs{y})$ is a smooth divergence measure.
\label{def:relaxed_primal}
\end{definition}
We may also relax only one of the marginal constraints:
\begin{definition}{Semi-relaxed smooth primal}
\begin{equation}
\normalfont
\OTphir(\bs{a}, \bs{b}) \coloneqq 
\min_{\substack{\T \ge 0\\ \T^\tr \bs{1}_m=\bs{b}}} ~
\langle \T, C \rangle 
+ \Phi(\T \bs{1}_n, \bs{a})
\label{eq:semi_relaxed_primal}
\end{equation}
where $\Phi(\bs{x}, \bs{y})$ is defined as in
Definition \ref{def:relaxed_primal}.
\end{definition}
For both
\eqref{eq:relaxed_primal} and \eqref{eq:semi_relaxed_primal}, the 
transportation plans will be typically sparse.
As discussed in more details in \S\ref{sec:related_work},
these formulations are similar to \citep{w_loss,chizat_unbalanced}, with the key
difference that we do not regularize $T$ with an entropic term.
In addition,
for $\Phi$, we propose to use
$\Phi(\bs{x}, \bs{y}) = \frac{1}{2\gamma} \|\bs{x} - \bs{y}\|^2$,
which is $\frac{1}{\gamma}$-smooth, while these works use a generalized 
Kullback-Leibler (KL)
divergence, which is not smooth.
Relaxing the marginal constraints is useful when normalizing input measures to
unit mass is not suitable \citep{agramfort} or to allow for only partial
displacement of mass.  Relaxing only one of the two constraints is useful in
color transfer \citep{adaptive_color_transfer}, where we would like all the
probability mass of the source image to be accounted for but not necessarily for
the reference image.  

\textbf{Dual interpretation.} As we show in Appendix
\ref{appendix:proof_strong_dual}, in the case $\Phi(\bs{x}, \bs{y}) =
\frac{1}{2\gamma} \|\bs{x} - \bs{y}\|^2$, the dual of \eqref{eq:relaxed_primal}
can be interpreted as the original dual with additional squared $2$-norm
regularization on the dual variables $\bs{\alpha}$ and $\bs{\beta}$.  For the
dual of \eqref{eq:semi_relaxed_primal}, the additional regularization is on
$\bs{\alpha}$ only (on the original dual or equivalently on the original
semi-dual). For that choice of $\Phi$, 
the duals of \eqref{eq:relaxed_primal} and
\eqref{eq:semi_relaxed_primal} are strongly convex.  The dual formulations are
crucial to derive our bounds in \S\ref{sec:bounds}.

\textbf{Solving the optimization problems.} While the relaxed and semi-relaxed
primals \eqref{eq:relaxed_primal} and \eqref{eq:semi_relaxed_primal} are still
constrained problems, it is much easier to project on their constraint domain
than on $\U(\va, \vb)$. For the relaxed primal, in our experiments we use L-BFGS-B, a variant of
L-BFGS suitable for box-constrained problems \citep{lbfgs_b}.
For the semi-relaxed primal, we use FISTA \citep{fista}.
Since the constraint domain of \eqref{eq:semi_relaxed_primal} has the structure
of a Cartesian product $b_1 \Simplex^m \times \dots \times b_n \Simplex^m$, we can
easily project any $\T$ on it by column-wise projection on the (scaled) simplex.
Although not exlored in this paper, the block Frank-Wolfe algorithm
\citep{block_fw} is also a good fit for the semi-relaxed primal.

\section{Theoretical bounds}
\label{sec:bounds}

\textbf{Convergence rates.} The dual \eqref{eq:smoothed_dual} is not smooth in
$\valpha$ and $\vbeta$ when using entropic regularization but it is when using
the squared $2$-norm, with constant upper-bounded by
$\nicefrac{n}{\gamma}$ \wrt $\valpha$ and $\nicefrac{m}{\gamma}$ \wrt $\vbeta$.
The semi-dual \eqref{eq:smoothed_semi_dual} is smooth for both
regularizations, with the same
constant of $\nicefrac{1}{\gamma}$, albeit not in the same norm.  The relaxed
and semi-relaxed primals \eqref{eq:relaxed_primal} and
\eqref{eq:semi_relaxed_primal} are both $\nicefrac{1}{\gamma}$-smooth
when using $\Phi(\bs{x}, \bs{y}) = \frac{1}{2\gamma} \|\bs{x} - \bs{y}\|^2$.
However, none of these problems are strongly convex. From standard convergence
analysis of (projected) gradient descent for smooth but non-strongly
convex problems, the number of iterations to reach an
$\epsilon$-accurate solution \wrt the smoothed problems is
$O(\nicefrac{1}{\gamma \epsilon})$ or $O(\nicefrac{1}{\sqrt{\gamma\epsilon}})$
with Nesterov acceleration. 
\begin{figure*}[t]
\centering
\includegraphics[scale=0.67]{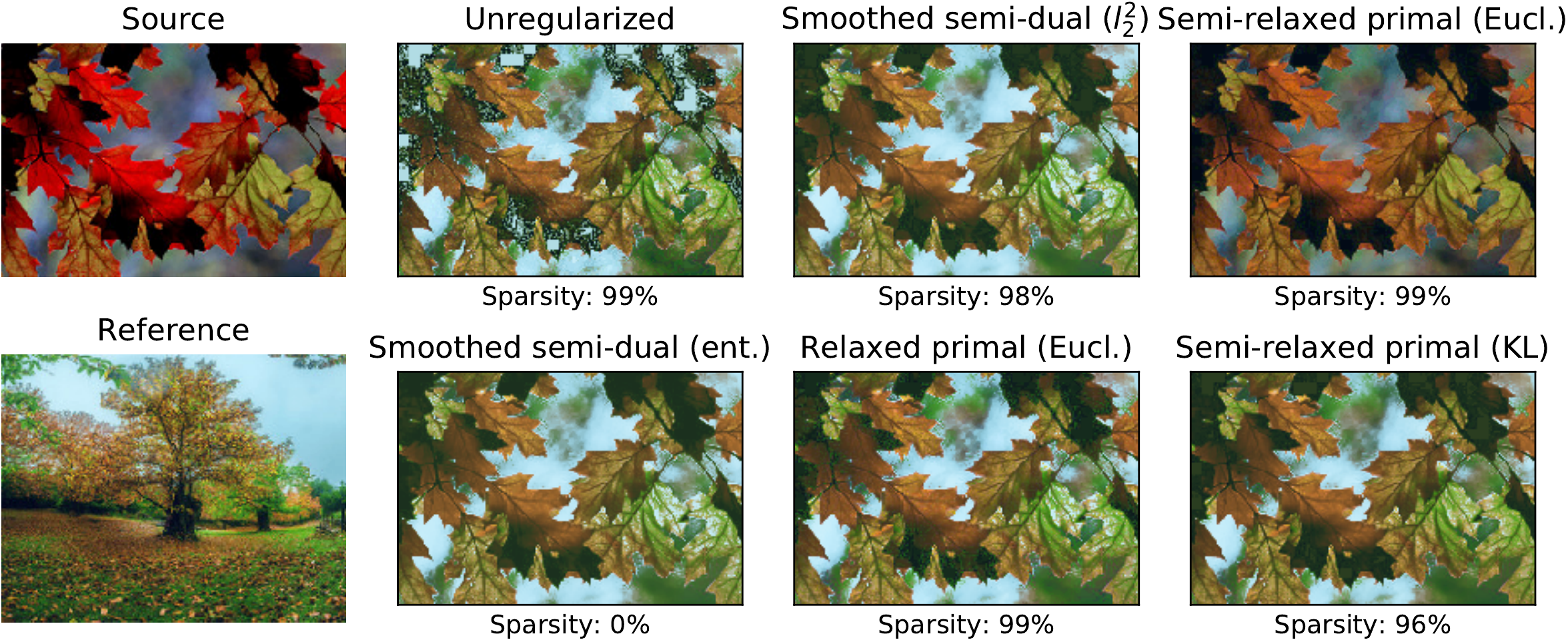}
\caption{Result comparison for different formulations on the task of color
transfer. For regularized formulations, we solve the optimization problem with 
$\gamma \in \{10^{-4},10^{-2},\dots,10^{4}\}$ and choose the most visually
pleasing result. The sparsity indicated below each image is the percentage of
zero elements in the optimal transportation plan.}
\label{fig:color_transfer_cmp}
\end{figure*}

\textbf{Approximation error.} Because the smoothed problems approach 
unregularized $\OT$ as $\gamma \to 0$, there is a trade-off between
convergence rate \wrt the smoothed problem and approximation error \wrt
unregularized $\OT$. A question is then which smoothed formulations and
which regularizations have better approximation error. Our first theorem bounds
$\OTomega - \OT$ in the case of entropic and squared $2$-norm
regularization.
\begin{theorem}{Approximation error of $\OTomega$}
    
Let $\bs{a} \in \Simplex^m$ and $\bs{b} \in \Simplex^n$. Then,
\begin{equation}
\normalfont
\gamma L
\le \OTomega(\va,\vb) - \OT(\va, \vb)
\le \gamma U,
\end{equation}
where we defined $L$ and $U$ as follows.

\begin{tabular}{c c c}

\toprule
$\Omega$ & \textnormal{Neg. entropy} & \textnormal{Squared $2$-norm} \\
\midrule
$L$ & $-H(\bs{a}) - H(\bs{b})$ & 
$\frac{1}{2} \displaystyle{\sum_{i,j=1}^{m,n}} \left(\frac{{a}_i}{n} + 
    \frac{{b}_j}{m} - \frac{1}{mn}\right)^2$ \\
$U$ & $-\max\{H(\bs{a}), H(\bs{b})\}$ & 
    $\frac{1}{2} \min\left\{ \|\bs{a}\|^2, \|\bs{b}\|^2\right\}$ \\
\bottomrule
\end{tabular}
\label{theorem:approx_error_OTomega}
\end{theorem}
Proof is given in Appendix \ref{appendix:proof_theorem_approx_error_OTomega}.
Our result suggests that, for the same $\gamma$, the approximation error can
often be smaller with squared $2$-norm than with entropic regularization. In
particular, this is true whenever $\min\{H(\va), H(\vb) \} > \frac{1}{2}
\min\left\{ \|\va\|^2, \|\vb\|^2\right\}$, which is often the case in
practice since  
$0 \le \min\{H(\va), H(\vb)\} \le \min\{\log m, \log n\}$
while
$0 \le \frac{1}{2} \min\left\{ \|\va\|^2, \|\vb\|^2\right\} \le \frac{1}{2}$.
Our second theorem bounds $\OT-\OTphi$ and $\OT-\OTphir$ when $\Phi$ is 
the squared Euclidean distance.
\begin{theorem}{Approximation error of $\OTphi$, $\OTphir$}
    
Let $\bs{a} \in \Simplex^m$, $\bs{b} \in \Simplex^n$,
$\Phi(\bs{x}, \bs{y}) = \frac{1}{2\gamma} \|\bs{x} - \bs{y}\|^2$.
Then,
\begin{align}
\normalfont
0 \le \OT(\bs{a}, \bs{b}) - \OTphi(\va, \vb) \le \gamma L \\
\normalfont
0 \le \OT(\bs{a}, \bs{b}) - \OTphir(\va, \vb) 
\le \gamma \widetilde{L} 
\end{align}
where we defined
\begin{equation}
\begin{split}
    L &\coloneqq \|C\|^2_\infty \min\{\nu_1 + n, \nu_2 + m\}^2 \\
    \nu_1 &\coloneqq \max \left\{ (2+n/m)\norminf{\va^{-1}},~
    \norminf{\vb^{-1}}\right\} \\
    \nu_2 &\coloneqq \max \left\{\norminf{\va^{-1}},~
    (2+m/n)\norminf{\vb^{-1}}\right\} \\
\widetilde{L} &\coloneqq 2 \norminf{C}^2 \norminf{\va^{-1}}^2.
\end{split}
\end{equation}
\label{theorem:approx_error_OTphi}
\end{theorem}
\vspace{-0.4cm}
Proof is given in Appendix \ref{appendix:proof_theorem_approx_error_OTphi}.
While the bound for $\OTphir$ is better than that of $\OTphi$, both are worse
than that of $\OTomega$, suggesting that the smoothed dual formulations are the way
to go when low approximation error \wrt unregularized $\OT$ is important.

\section{Experimental results}
\label{sec:exp}

We showcase our formulations on color transfer, which is a classical OT application
\citep{color_transfer}. 
More experimental results are presented in Appendix \ref{appendix:exp}.

\subsection{Application to color transfer}
\label{sec:app_color_transfer}
\begin{figure*}[t]
\centering
\includegraphics[scale=0.38]{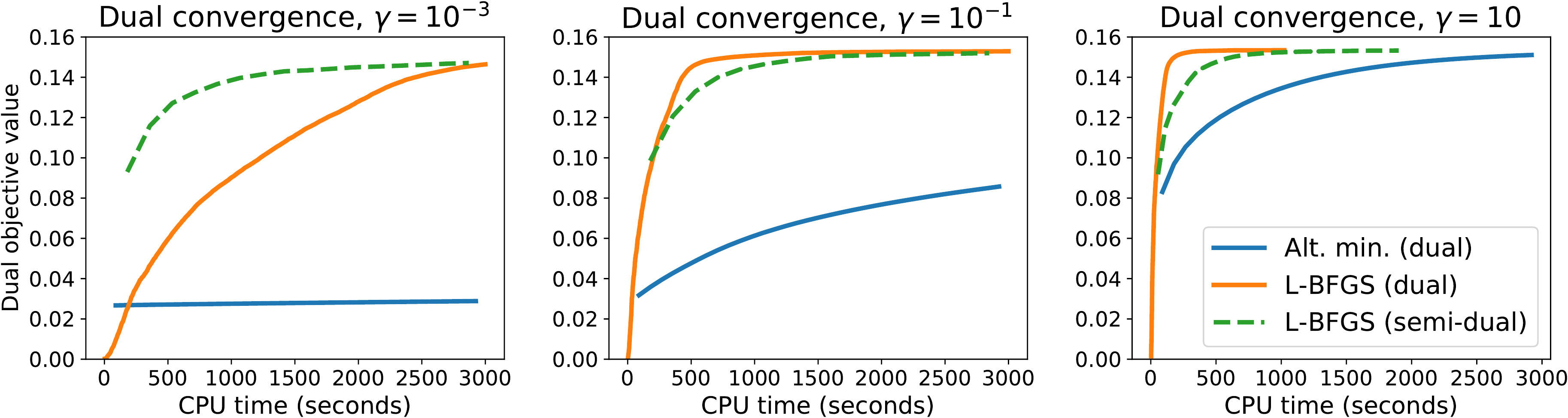}
\caption{Solver comparison for the smoothed dual and semi-dual, with
squared $2$-norm regularization. With $\gamma=10$, which was also the best value
selected in Figure \ref{fig:color_transfer_cmp}, the maximum is reached in less
than 4 minutes.}
\label{fig:convergence}
\end{figure*}

\textbf{Experimental setup.} Given an image of size $u \times v$, we represent
its pixels in RGB color space. We apply k-means clustering to quantize the
image down to $m$ colors. This produces $m$ color centroids
$\bs{x}_1, \dots, \bs{x}_m \in \mathbb{R}^3$. We can
count how many pixels were assigned to each centroid and normalizing by $uv$
gives us a color histogram $\va \in \Simplex^m$. We repeat the same process
with a second image to obtain $\bs{y}_1,\dots,\bs{y}_n \in \mathbb{R}^3$ and
$\vb \in \Simplex^n$. Next, we apply any of the proposed methods with cost matrix
$\matelem{C}{i}{j} = d(\bs{x}_i, \bs{y}_j)$, where $d$ is some discrepancy
measure, to obtain a (possibly relaxed) transportation plan
$\T \in \mathbb{R}_+^{m \times n}$. For each color centroid
$\bs{x}_i$, we apply a barycentric projection to obtain a new color centroid
\begin{equation}
\hat{\bs{x}}_i \coloneqq 
\argmin_{\bs{x} \in \mathbb{R}^3} \sum_{j=1}^n \matelem{T}{i}{j} ~ d(\bs{x},
\bs{y}_j).
\end{equation}
When $d(\bs{x}, \bs{y}) = \|\bs{x} - \bs{y}\|^2$, as used in our experiments,
the above admits a closed-form solution: $\hat{\bs{x}}_i = \frac{\sum_{j=1}^n
\matelem{t}{i}{j} \bs{y}_j}{\sum_{j=1}^n \matelem{T}{i}{j}}$. Finally, we use
the new color $\hat{\bs{x}}_i$ for all pixels assigned to $\bs{x}_i$. The same
process can be performed with respect to the $\bs{y}_j$, in order to transfer
the colors in the other direction. 
We use two public domain images ``fall foliage''
by Bernard Spragg and ``comunion'' by Abel Maestro Garcia, and reduce the number
of colors to $m=n=4096$. 
We compare smoothed dual approaches and (semi-)relaxed primal approaches. 
For the semi-relaxed primal, we also compared with $\Phi(\vx, \vy) =
\frac{1}{\gamma} \KL(\vx||\vy)$, where $\KL(\vx||\vy)$ is the generalized KL
divergence, $\vx^\tr \log\left(\frac{\vx}{\vy}\right) - \vx^\tr \bs{1} + \vy^\tr
\bs{1}$. This choice is differentiable but not smooth.
We ran the aforementioned solvers for up to $1000$ epochs.

\textbf{Results.} Our results are presented in Figure
\ref{fig:color_transfer_cmp}.  
All formulations clearly produced better
results than unregularized OT.  
With the exception of the entropy-smoothed semi-dual
formulation, all formulations produced extremely sparse transportation
plans. The semi-relaxed primal formulation with $\Phi$ set to the squared
Euclidean distance was the only one to produce colors with a darker tone.

\subsection{Solver and objective comparison}
\label{sec:solver_cmp}

We compared the smoothed dual and semi-dual when using squared
$2$-norm regularization. In addition to L-BFGS on both objectives, we also
compared with alternating minimization in the dual. As we show in Appendix
\ref{appendix:alt_min}, exact block minimization \wrt $\valpha$ and $\vbeta$ can
be carried out by projection onto the simplex.

\textbf{Results.}
We ran the comparison using the same data as in
\S\ref{sec:app_color_transfer}. Results are indicated in Figure
\ref{fig:convergence}. When the problem is loosely regularized, we made two key
findings: i) L-BFGS converges much faster in the semi-dual
than in the dual, ii) alternating minimization converges extremely slowly.
The reason for i) could be the better smoothness constant of the semi-dual
(cf.\ \S\ref{sec:bounds}). Since alternating minimization and the semi-dual have
roughly the same cost per iteration (cf. Appendix \ref{appendix:alt_min}), the
reason for ii) is \textit{not} iteration cost but a convergence issue of alternating
minimization. When using larger regularization, L-BFGS appears to converge
slighly faster on the dual than on the semi-dual, which is likely thanks to its
cheap-to-compute gradients.

\subsection{Approximation error comparison}

\begin{figure*}[t]
\centering
\includegraphics[scale=0.51]{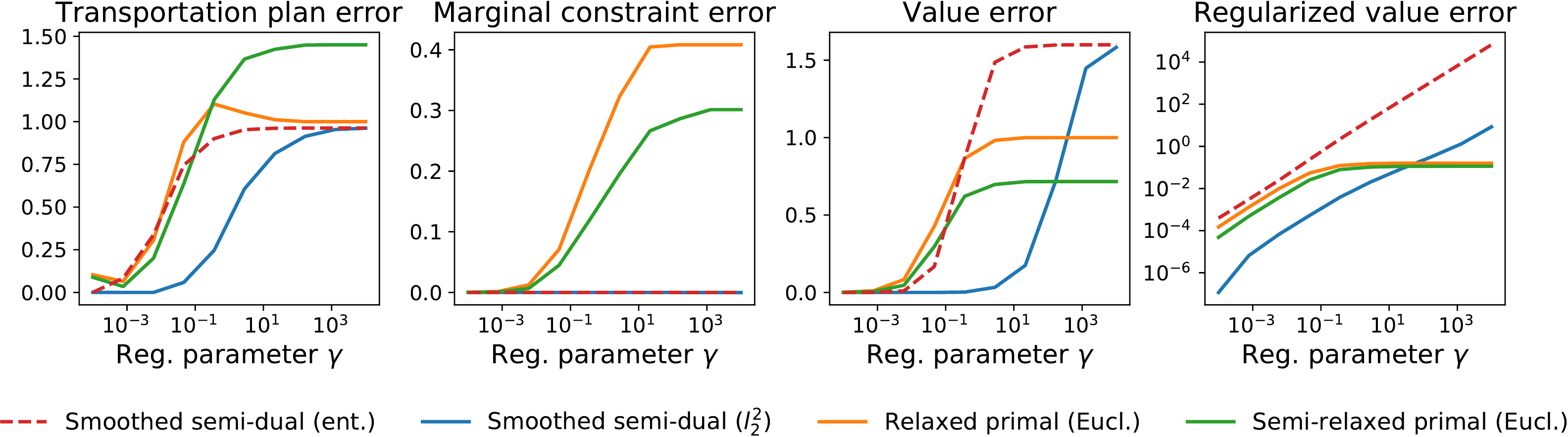}
\caption{Approximation error \wrt unregularized OT empirically achieved by
    different smoothed formulations on the task of color transfer. 
Let $\T^\star$ be en optimal solution of the unregularized LP
\eqref{eq:primal_lp} and $T^\star_\gamma$
be an optimal solution of one of the smoothed formulations with
regularization parameter $\gamma$. The transportation plan error is 
$\|\T^\star_\gamma-\T^\star\| / \|\T^\star\|$. The marginal constraint error is
$\|T^\star_\gamma \bs{1}_n - \va\| + \|(T^\star_\gamma)^\tr \bs{1}_m - \vb\|$.
The value error is $|\langle T^\star_\gamma, C \rangle - \langle T^\star, C
\rangle| / \langle T^\star, C \rangle$.
The regularized value error is $|v - \langle T^\star, C \rangle|$, where $v$ is
one of $\OTomega(\va, \vb)$, $\OTphi(\va, \vb)$ and $\OTphir(\va, \vb)$.
For the regularized value error, our empirical findings confirm what Theorem
\ref{theorem:approx_error_OTomega} suggested, namely that, for the same
value of $\gamma$, squared $2$-norm regularization is quite tighter than
entropic regularization.
}
\label{fig:approx_error}
\end{figure*}

We compared empirically the approximation error of smoothed formulations 
\wrt unregularized OT according to four criteria: transportation
plan error, marginal constraint error, value error and regularized value error
(cf.  Figure \ref{fig:approx_error} for a precise definition). For the 
dual approaches, we solved the smoothed semi-dual objective
\eqref{eq:smoothed_semi_dual}, since, as we discussed in \S\ref{sec:bounds}, it
has the same smoothness constant of $\nicefrac{1}{\gamma}$ for both
entropic and squared $2$-norm regularizations, implying similar convergence
rates in theory. In addition, in the case of entropic regularization, the
expressions of $\maxop$ and $\nabla \maxop$ are trivial to stabilize numerically
using standard log-sum-exp implementation tricks. 

\textbf{Results.} We ran the comparison using the same data as in
\S\ref{sec:app_color_transfer}. Results are indicated in Figure
\ref{fig:approx_error}. For the transportation plan error and the (regularized)
value error, entropic regularization required 100 times smaller $\gamma$ to
achieve the same error. This confirms, as suggested by Theorem
\ref{theorem:approx_error_OTomega}, that squared $2$-norm regularization is
typically tighter. Unsurprisingly, the semi-relaxed primal was tighter than
the relaxed primal in all four criteria. A runtime comparison of
smoothed formulations is also important. However, a rigorous comparison would
require carefully engineered implementations and is therefore left for future
work.

\section{Related work}
\label{sec:related_work}

\textbf{Regularized OT.} Problems similar to \eqref{eq:regularized_primal} for
general $\Omega$ were considered in \citep{rot_mover}.  Their work focuses on
strictly convex and differentiable $\Omega$ for which there exists an associated
Bregman divergence. Following \citep{iterative_bregman_proj}, they show that
\eqref{eq:regularized_primal} can then be reformulated as a Bregman projection
onto the transportation polytope and solved using Dykstra's algorithm
[\citeyear{dykstra_algo}]. While Dykstra's algorithm can be interpreted
implicitly as a two-block alternating minimization scheme on the dual problem,
neither the dual nor the semi-dual expressions were derived. These expressions
allow us to make use of arbitrary solvers, including quasi-Newton ones like
L-BFGS, which as we showed empirically, converge much faster on loosely
regularized problems. Our framework can also accomodate non-differentiable
regularizations for which there does not exist an associated Bregman divergence,
such as those that include a group lasso term.  Squared $2$-norm regularization
was recently considered in \citep{li2016fast} as well as in
\citep{quadratically_reg_OT} but for a reformulation of the Wasserstein distance
of order $1$ as a min cost flow problem along the edges of a graph. 

\textbf{Relaxed OT.} There has been a large number of proposals to extend OT to
unbalanced positive measures.  Static formulations with approximate marginal
constraints based on the KL divergence have been proposed in
\citep{w_loss,chizat_unbalanced}.  The main difference with our work is that
these formulations include an additional entropic regularization on $\T$. While
this entropic term enables a Sinkhorn-like algorithm, it also prevents from
obtaining sparse $\T$ and requires the tuning of an additional hyper-parameter.
Relaxing only one of the two marginal constraints with an inequality was
investigated for color transfer in \citep{adaptive_color_transfer}.
\citet{benamou_unbalanced} considered an interpolation between OT and squared
Euclidean distances:
\begin{equation}
\min_{\bs{x} \in \Simplex^m} 
\text{OT}(\bs{x}, \bs{b}) +
\frac{1}{2\gamma} \|\bs{x} - \bs{a}\|^2.
\label{eq:OT_euc_interpolation}
\end{equation}
While on first sight this looks quite different, this is in fact equivalent to
our semi-relaxed primal formulation when
$\Phi(\bs{x}, \bs{y}) = \frac{1}{2\gamma} \|\bs{x} - \bs{y}\|^2$ since
\eqref{eq:OT_euc_interpolation} is equal to
\begin{align}
&\min_{\bs{x} \in \Simplex^m} 
\min_{\substack{\T \ge 0\\ \T \bs{1}_n=\bs{x}\\ \T^\tr \bs{1}_m=\vb}}
\langle \T, C \rangle + \frac{1}{2\gamma} \|\bs{x} - \bs{a}\|^2 \\
=& \min_{\substack{\T \ge 0\\ \T^\tr \bs{1}_m=\vb}}
\langle \T, C \rangle + \frac{1}{2\gamma} \|\T \bs{1}_n - \bs{a}\|^2
= \OTphir(\va, \vb).
\end{align}
However, the bounds in \S\ref{sec:bounds} are to our knowledge new.
A similar formulation but with a group-lasso penalty on $T$ instead of
$\frac{1}{2\gamma} \|\T \bs{1}_n - \bs{a}\|^2$ was considered in the context of
convex clustering \citep{carli2013convex}.

\textbf{Smoothed LPs.} Smoothed linear programs have been investigated in other
contexts. The two closest works to ours are \citep{ssvm_soft} and
\citep{smooth_and_strong}, in which smoothed LP relaxations based on the squared
$2$-norm are proposed for maximum a-posteriori inference.  One innovation we
make compared to these works is to abstract away the regularization by
introducing the $\indomega$ and $\maxop$ functions.

\section{Conclusion}

We proposed in this paper to regularize both the primal and dual OT
formulations with a
strongly convex term, and showed that this corresponds to relaxing the dual and
primal constraints with smooth approximations.
There are several important avenues for future work. The conjugate expression
\eqref{eq:ot_omega_conjugate} should be useful for barycenter computation
\citep{variational_cuturi} or dictionary learning \citep{paper_antoine} with
squared $2$-norm instead of entropic regularization. On the theoretical side, 
while we provided convergence guarantees \wrt the OT distance \textit{value} as
the regularization vanishes,
which suggested the advantage of squared $2$-norm regularization,
it would also be important to study the convergence \wrt the 
\textit{transportation plan}, as was done for entropic regularization by
\citet{asymptotic_analysis_trajectory}. Finally, studying optimization
algorithms that can cope with large-scale data is important. We believe SAGA
\citep{saga} is a good candidate since
it is stochastic, supports proximity operators, is adaptive to non-strongly
convex problems and can be parallelized \citep{asaga}.

\clearpage

\section*{Acknowledgements}

We thank Arthur Mensch and the anonymous
reviewers for constructive comments.


\clearpage
\onecolumn
\appendix

\begin{center}
    {\Huge \bf Appendix}
\end{center}

\section{Proofs}

\subsection{Derivation of the smooth relaxed dual}
\label{appendix:proof_dual}

Recall that
\begin{equation}
    \OTomega(\bs{a}, \bs{b}) = \min_{\T \in \mathcal{U}(\bs{a},\bs{b})}
\sum_{j=1}^n \col{\T}{j}^\tr \col{c}{j} + \Omega(\col{\T}{j}).
\label{eq:regularized_primal_bis}
\end{equation}

We now add Lagrange multipliers for the two equality constraints but keep
the constraint $\T \ge 0$ explicitly:
\begin{equation}
\OTomega(\bs{a}, \bs{b}) = 
\min_{\T \ge 0} 
\max_{\bs{\alpha} \in \mathbb{R}^m, \bs{\beta} \in \mathbb{R}^n} 
\sum_{j=1}^n \col{\T}{j}^\tr \col{c}{j} + 
\Omega(\col{\T}{j}) +
\bs{\alpha}^\tr(\T \bs{1}_n - \bs{a}) + \bs{\beta}^\tr(\T^\tr \bs{1}_m -
\bs{b}).
\end{equation}
Since \eqref{eq:regularized_primal_bis} is a convex optimization problem 
with only linear equality and inequality constraints, Slater's conditions reduce
to feasibility \citep[\S5.2.3]{boyd_book} and hence strong duality holds:
\begin{align}
\OTomega(\bs{a}, \bs{b}) &= 
\max_{\bs{\alpha} \in \mathbb{R}^m, \bs{\beta} \in \mathbb{R}^n} 
\min_{\T \ge 0} \sum_{j=1}^n \col{\T}{j}^\tr \col{c}{j} + 
\Omega(\col{\T}{j}) +
\bs{\alpha}^\tr(\T \bs{1}_n - \bs{a}) + \bs{\beta}^\tr(\T^\tr \bs{1}_m -
\bs{b}) \\
&=\max_{\bs{\alpha} \in \mathbb{R}^m, \bs{\beta} \in \mathbb{R}^n} 
\sum_{j=1}^n \min_{\col{\T}{j} \ge 0} \col{\T}{j}^\tr (\col{c}{j} + \bs{\alpha}
+ \beta_j \bs{1}_m) + \Omega(\col{\T}{j}) -\bs{\alpha}^\tr \bs{a}-
\bs{\beta}^\tr \bs{b} \\
&=\max_{\bs{\alpha} \in \mathbb{R}^m, \bs{\beta} \in \mathbb{R}^n} 
-\sum_{j=1}^n \max_{\col{\T}{j} \ge 0} \col{\T}{j}^\tr (-\col{c}{j} - \bs{\alpha}
- \beta_j \bs{1}_m) - \Omega(\col{\T}{j}) -\bs{\alpha}^\tr \bs{a}-
\bs{\beta}^\tr \bs{b} \\
&=\max_{\bs{\alpha} \in \mathbb{R}^m, \bs{\beta} \in \mathbb{R}^n} 
\bs{\alpha}^\tr \bs{a}+ \bs{\beta}^\tr \bs{b} 
-\sum_{j=1}^n \max_{\col{\T}{j} \ge 0}
\col{\T}{j}^\tr (\bs{\alpha} + \beta_j \bs{1}_m - \col{c}{j}) - 
\Omega(\col{\T}{j}).
\end{align}
Finally, plugging the expression of \eqref{eq:smoothed_indicator} gives the
claimed result.

\subsection{Derivation of the convex conjugate}
\label{appendix:proof_conjugate}

The convex conjugate of $\OTomega(\bs{a},\bs{b})$ \wrt the first argument is
\begin{equation}
\OTomega^*(\bs{g}, \bs{b}) = \sup_{\bs{a} \in \Simplex^m} \bs{g}^\tr \bs{a} - 
\OTomega(\bs{a}, \bs{b}).
\end{equation}

Following a similar argument as \citep[Theorem 2.4]{variational_cuturi}, we have
\begin{equation}
\OTomega^*(\bs{g}, \bs{b}) = 
\max_{\substack{\T \ge 0\\ \T^\tr \bs{1}_m = \bs{b}}}
\langle \T, \bs{g} \bs{1}_n^\tr - C \rangle 
- \sum_{j=1}^n \Omega(\col{\T}{j}).
\label{eq:conjugate_intermediate}
\end{equation}
Notice that this is an easier optimization problem than
\eqref{eq:regularized_primal}, since there are equality constraints only in one
direction.  \citet{variational_cuturi} showed that this optimization problem
admits a closed form in the case of entropic regularization. Here, we show how
to compute $\OTomega^*$ for any strongly-convex regularization. 

The problem clearly decomposes over columns and we can rewrite it as
\begin{align}
\OTomega^*(\bs{g}, \bs{b}) 
&= \sum_{j=1}^n 
\max_{\substack{\col{\T}{j} \ge 0\\
\col{\T}{j}^\tr \bs{1}_m = b_j}}
\col{\T}{j}^\tr (\bs{g} - \col{c}{j}) - \Omega(\col{\T}{j}) \\
&= \sum_{j=1}^n 
b_j \max_{\col{\tau}{j} \in \Simplex^m}
\col{\tau}{j}^\tr (\bs{g} - \col{c}{j}) - \frac{1}{b_j} 
\Omega(b_j \col{\tau}{j}) \\
&= \sum_{j=1}^n b_j \maxopj(\bs{g} - \col{c}{j}),
\end{align}
where we defined $\Omega_j(\bs{y}) \coloneqq \frac{1}{b_j} \Omega(b_j \bs{y})$
and where $\maxop$ is defined in \eqref{eq:maxop}.

\subsection{Expression of the strongly-convex duals}
\label{appendix:proof_strong_dual}

Using a similar derivation as before, we obtain the duals of
\eqref{eq:relaxed_primal} and \eqref{eq:semi_relaxed_primal}.
\begin{proposition}{Duals of \eqref{eq:relaxed_primal} and
    \eqref{eq:semi_relaxed_primal}}
    
\begin{align}
\normalfont
\OTphi(\va, \vb) &= \max_{\valpha, \vbeta \in \mathcal{P}(C)}
-\frac{1}{2}\Phi^*(-2\valpha, \va) -\frac{1}{2}\Phi^*(-2\vbeta, \vb) \\
\normalfont
\OTphir(\va, \vb)  
&= \max_{\valpha, \vbeta \in \mathcal{P}(C)}
-\Phi^*(-\valpha, \va) + \vbeta^\tr \vb \\
&=\max_{\valpha \in \mathbb{R}^m} 
-\Phi^*(-\valpha, \va)
- \sum_{j=1}^n b_j \max_{i \in [m]}(\alpha_i - \matelem{C}{i}{j}),
\end{align}
where $\Phi^*$ is the conjugate of $\Phi$ in the first argument.
\end{proposition}
The duals are strongly convex if $\Phi$ is smooth. \\
When $\Phi(\bs{x}, \bs{y}) = \frac{1}{2\gamma} \|\bs{x} - \bs{y}\|^2$,
$\Phi^*(-\valpha, \va) = \frac{\gamma}{2} \|\valpha\|^2 - \valpha^\tr
\va$. Plugging that expression in the above, we get
\begin{equation}
\OTphi(\va, \vb)
= \max_{\valpha, \vbeta \in \mathcal{P}(C)}
\valpha^\tr \va + \vbeta^\tr \vb - \gamma \left( \|\valpha\|^2 +
\|\vbeta\|^2 \right)
\label{eq:strongly_convex_dual_l2}
\end{equation}
and
\begin{align}
\OTphir(\va, \vb) 
&= \max_{\valpha, \vbeta \in \mathcal{P}(C)}
\valpha^\tr \va + \vbeta^\tr \vb - \frac{\gamma}{2} \|\valpha\|^2 \\
&=\max_{\valpha \in \mathbb{R}^m} 
\valpha^\tr \va 
- \sum_{j=1}^n b_j \max_{i \in [m]}(\alpha_i - \matelem{C}{i}{j})
- \frac{\gamma}{2} \|\valpha\|^2.
\label{eq:strongly_convex_semi_dual_l2}
\end{align}
This corresponds to the original dual and semi-dual with squared $2$-norm
regularization on the variables.

\subsection{Proof of Theorem \ref{theorem:approx_error_OTomega}}
\label{appendix:proof_theorem_approx_error_OTomega}

Before proving the theorem, we introduce the next two lemmas, which bound the
regularization value achieved by any transportation
plan.

\begin{lemma}{Bounding the entropy of a transportation plan}

Let $H(\bs{a}) \coloneqq -\sum_i a_i \log a_i$ and $H(\T)
\coloneqq -\sum_{i,j} \matelem{\T}{i}{j} \log \matelem{\T}{i}{j}$ be the
joint entropy.  \\
Let $\bs{a} \in \Simplex^m$, $\bs{b} \in \Simplex^n$ and $\T \in
\mathcal{U}(\bs{a}, \bs{b})$.  Then,
\begin{equation}
\max\{H(\va), H(\vb)\} \le H(\T) \le H(\bs{a}) + H(\bs{b}).
\end{equation}
\label{lemma:bound_entropy}
\end{lemma}
\textit{Proof.} See, for instance, \citep{cover_and_thomas}.

Together with $0 \le H(\bs{a}) \le \log m$
and $0 \le H(\bs{b}) \le \log n$, this provides lower and upper bounds for the
entropy of a transportation plan. As noted in \citep{sinkhorn_distances}, the
upper bound is tight since
\begin{equation}
\max_{\T \in \U(\va,\vb)} H(\T) = H(\va\vb^\tr) = H(\va) + H(\vb).
\end{equation}

\begin{lemma}{Bounding the squared $2$-norm of a transportation plan}

Let $\bs{a} \in \Simplex^m$, $\bs{b} \in \Simplex^n$ and $\T \in
\mathcal{U}(\bs{a}, \bs{b})$.  Then,
\begin{equation}
\sum_{i=1}^m \sum_{j=1}^{n} \left(\frac{{a}_i}{n} + \frac{{b}_j}{m} - 
\frac{1}{mn}\right)^2 \le 
\|\T\|^2 \le 
\min\left\{ \|\bs{a}\|^2, \|\bs{b}\|^2\right\}.
\end{equation}
\label{lemma:bound_squared_l2}
\end{lemma}
\textit{Proof.} The tightest lower bound is given by $\displaystyle{\min_{\T \in
\U(\va, \vb)}} \|T\|^2$. An exact iterative algorithm
was proposed in \citep{closest_point_tp} to solve this problem. However, since
we are interested in an explicit formula, we consider instead the lower bound
$\displaystyle{\min_{\substack{\T \bs{1}_n = \va\\\T^\tr \bs{1}_m = \vb}}}
\|T\|^2$ (\textit{i.e}., we ignore the non-negativity constraint). It is known
\citep{easy_tp_like} that the minimum is achieved at $\matelem{t}{i}{j} =
\frac{{a}_i}{n} + \frac{{b}_j}{m} - \frac{1}{mn}$, hence our lower bound.
For the upper bound, we have
\begin{align}
\|\T\|^2
&= \sum_{i=1}^m \sum_{j=1}^n \matelem{\T}{i}{j}^2 \\
&= \sum_{i=1}^m \sum_{j=1}^n \left(a_i \frac{\matelem{\T}{i}{j}}{a_i}\right)^2 \\
&= \sum_{i=1}^m a_i^2 \sum_{j=1}^n \left(\frac{\matelem{\T}{i}{j}}{a_i}\right)^2 \\
&\le \sum_{i=1}^m a_i^2 \sum_{j=1}^n \left(\frac{\matelem{\T}{i}{j}}{a_i}\right) \\
&= \|\bs{a}\|^2.
\end{align}
We can do the same with $\bs{b} \in \Simplex^n$ to obtain $\|\T\|^2 \le
\|\bs{b}\|^2$, yielding the claimed result.
$\square$

Together with $0 \le \|\va\|^2 \le 1$ and $0 \le \|\vb\|^2 \le 1$, this provides
lower and upper bounds for the squared $2$-norm of a transportation plan.

\textbf{Proof of the theorem.} Let $\T^\star$ and $\T^\star_\Omega$ be
optimal solutions of \eqref{eq:primal_lp} and \eqref{eq:regularized_primal},
respectively. Then,
\begin{equation}
\OT(\bs{a}, \bs{b}) + \Omega(\T^\star_\Omega)
= \langle \T^\star, C \rangle + \Omega(\T^\star_\Omega) 
\le \langle \T^\star_\Omega, C \rangle + \Omega(\T^\star_\Omega) 
= \OTomega(\bs{a}, \bs{b}).
\end{equation}
Likewise,
\begin{equation}
\OTomega(\bs{a}, \bs{b}) 
= \langle \T^\star_\Omega, C \rangle + \Omega(\T^\star_\Omega) 
\le \langle \T^\star, C \rangle + \Omega(\T^\star) 
= \OT(\bs{a}, \bs{b}) + \Omega(\T^\star).
\end{equation}
Combining the two, we obtain
\begin{equation}
\OT(\bs{a}, \bs{b}) + \Omega(\T^\star_\Omega)
\le \OTomega(\bs{a}, \bs{b})
\le \OT(\bs{a}, \bs{b}) + \Omega(\T^\star).
\end{equation}
Using $\T^\star, \T^\star_\Omega \in \mathcal{U}(\bs{a}, \bs{b})$
together with Lemma \ref{lemma:bound_entropy} and Lemma
\ref{lemma:bound_squared_l2} gives the claimed results.

\subsection{Proof of Theorem \ref{theorem:approx_error_OTphi}}
\label{appendix:proof_theorem_approx_error_OTphi}

To prove the theorem, we first need the following two lemmas.

\begin{lemma}{Bounding the $1$-norm of $\valpha$ and $\vbeta$ for
$(\valpha,\vbeta) \in \mathcal{P}(C)$}
    
Let $\valpha,\vbeta \in \mathcal{P}(C)$ with extra constraints
$\valpha^\tr \bs{1}_m = 0$ and $\valpha^\tr \va + \vbeta^\tr \vb \ge 0$,
where $\bs{a} \in \Simplex^m$ and $\bs{b} \in \Simplex^n$.
Then,

\begin{equation}
0 \le \|\valpha\|_1 + \|\vbeta\|_1 \le \|C\|_\infty (\nu + n)
\end{equation}
where
\begin{equation}
\nu = \max \left\{ (2+n/m)\norminf{\va^{-1}},~ \norminf{\vb^{-1}}\right\}.
\end{equation}
\label{lemma:bound_1norm_PC}
\end{lemma}
\textit{Proof.} The proof technique is inspired by \citep[Supplementary material
Lemma 1.2]{convergence_rate_map}.

The 1-norm can be rewritten as
\begin{equation}
\normone{\valpha} + \normone{\vbeta} = 
\underset{\substack{\bs{r} \in \{-1,1\}^m\\
\bs{s} \in \{-1,1\}^n}}{\max} \quad 
\bs{r}^\tr \valpha + \bs{s}^\tr \vbeta.
\end{equation}
Our goal is to upper bound the following objective
\begin{equation}
\begin{split}
\underset{\valpha \in \mathbb{R}^m, \vbeta \in \mathbb{R}^n}{\max} \quad 
\bs{r}^\tr \valpha + \bs{s}^\tr \vbeta 
\quad \text{s.t.} \quad & 0 \le \valpha^\tr \va + \vbeta^\tr \vb, \\
& \alpha_i + \beta_j \le \matelem{C}{i}{j}, \\
& \valpha^\tr \bs{1}_m = 0,
\end{split}
\end{equation}
with a constant that does not depend on $\bs{r}$ and $\bs{s}$. 
We call the above the dual problem.
Its Lagrangian is
\begin{equation}
\begin{split}
L(\valpha, \vbeta, \mu, \nu, T) &=  
\bs{r}^\tr \valpha + \bs{s}^\tr \vbeta +
\mu \valpha^\tr \bs{1}_m + \nu(\valpha^\tr \va + \vbeta^\tr \vb) +
\sum_{i,j=1}^{m,n} \matelem{T}{i}{j} 
\left(\matelem{C}{i}{j} - \alpha_i - \beta_j\right)  \\
&= (\bs{r} + \mu \bs{1}_m + \nu \va - \T \bs{1}_n)^\tr \valpha
+ (\bs{s} + \nu \vb - \T^\tr \bs{1}_m)^\tr \vbeta
+ \langle \T, C \rangle
\end{split}
\end{equation}
with $\mu \in \mathbb{R}$, $\nu \ge 0$, $T \ge \bs{0}$. 
Maximizing the Lagrangian \wrt $\valpha$ and $\vbeta$ gives
the corresponding primal problem
\begin{equation}
\begin{split}
\min_{\T \ge 0, ~\mu \in \mathbb{R}, ~\nu \ge 0} 
~ \langle \T, C \rangle 
\quad \text{s.t.} \quad
&T\mathbf{1}_n = \nu \va + \bs{r} + \mu \mathbf{1}_m, \\
&T^\tr \mathbf{1}_m = \nu \vb + \bs{s}.
\end{split}
\end{equation}
By weak duality, any feasible primal point provides an upper bound of the dual
problem.  We start by choosing $\mu = \frac{1}{m}(\sum_j s_j- \sum_i r_i)$ so
that $\sum_{i,j} \matelem{T}{i}{j}$ provides the same values \wrt the last two
constraints. Next, we choose
\begin{equation}
\nu = \max \left\{  \underset{i}{\max} \frac{2+n/m}{a_i}, \underset{j}{\max}
\frac{1}{b_j}\right\}
\end{equation}
which ensures the non-negativity of $\nu\va + \bs{r} + \mu \mathbf{1}_m$ and
$\nu\vb + \bs{s}$ regardless of $\bs{r}$ and $\bs{s}$. It follows that the
transportation plan $T$ defined by
\begin{equation}
T = \frac{1}{(\nu \vb + \bs{s})^T\mathbf{1}_n}(\nu \va + \bs{r} + \mu
\mathbf{1}_m)(\nu \vb + \bs{s})^\tr
\end{equation}
is feasible. We finally bound the objective, $\langle T, C \rangle \le
\norminf{C} \sum_{i,j} \matelem{T}{i}{j} \le \norminf{C} (\nu+n)$. $\square$

\begin{lemma}{Bounding the $1$-norm of $\valpha$ for $(\valpha,\cdot) \in
\mathcal{P}(C)$}
    
Let $\valpha,\vbeta \in \mathcal{P}(C)$ with extra constraints
$\sum_{i=1}^m \alpha_i = 0$ and $\alpha^\tr \va + \vbeta^\tr \vb \ge 0$,
where $\bs{a} \in \Simplex^m$ and $\bs{b} \in \Simplex^n$.
Then,
\begin{equation}
    0 \le \|\valpha\|_1 \le 2\|C\|_\infty \norminf{\va^{-1}}.
\end{equation}
\label{lemma:bound_1norm_PC_semi}
\end{lemma}
\textit{Proof.} Similarly as before, our goal is to upper bound
\begin{equation}
\begin{split}
    \underset{\valpha \in \mathbb{R}^m,\vbeta \in \mathbb{R}^n}{\max} ~
\bs{r}^\tr \valpha
\quad \text{s.t.} \quad
&0 \le \valpha^\tr \va + \vbeta^\tr \vb, \\
& \alpha_i + \beta_j \le \matelem{C}{i}{j}, \\
& \valpha^\tr \bs{1}_m = 0,
\end{split}
\end{equation}
with a constant which does not depend on $\bs{r}$. 
The corresponding primal is
\begin{equation}
\begin{split}
\underset{\T \ge 0, ~\mu \in \mathbb{R}, ~\nu \ge 0}{\min} ~ \langle \T, C \rangle 
\quad \text{s.t.} \quad
& T\mathbf{1}_n = \nu \va + \bs{r} + \mu \mathbf{1}_m, \\
& T^\tr\mathbf{1}_m = \nu \vb.
\end{split}
\end{equation}
By weak duality, any feasible primal point gives us an upper bound.  We start by
choosing $\mu = \frac{1}{m}\sum_i r_i$ so that $\sum_{ij} \matelem{t}{i}{j}$
provides the same values \wrt the last two constraints. Next, we choose,
$\nu = \underset{i}{\max} \frac{2}{a_i}$,
which ensures the non-negativity of $\nu \va + \bs{r} + \mu \mathbf{1}_m$ ($\nu
\vb \ge 0$ is also satisfied since $\nu \ge 0$) which appears in the r.h.s.\ of
the second constraint, independently of $\bs{r}$. It follows that the
transportation plan $T$ defined by
\begin{equation}
    T = \frac{1}{\nu \vb^\tr \mathbf{1}_n}(\nu \va + \bs{r} + \mu
    \mathbf{1}_m)(\nu \vb)^\tr = (\nu \va + \bs{r} + \mu \mathbf{1}_m)\vb^\tr
\end{equation}
is feasible. We finally bound the objective
\begin{equation}
    \langle \T, C \rangle \le \norminf{C} \sum_{i,j} \matelem{T}{i}{j} \le \nu
    \norminf{C} = 2\norminf{C} \norminf{\va^{-1}},
\end{equation}
which concludes the proof. $\square$

\textbf{Proof of the theorem.} We begin by deriving the bound for the relaxed
primal. Let $(\valpha^\star,\vbeta^\star)$ and
$(\valpha^\star_\Phi,\vbeta^\star_\Phi)$
be optimal solutions of \eqref{eq:dual_lp} and
\eqref{eq:strongly_convex_dual_l2}, respectively. 
Since
$(\bs{\alpha}^\star_\Phi)^\tr \bs{a} + (\bs{\beta}^\star_\Phi)^\tr \bs{b} 
\le
(\bs{\alpha}^\star)^\tr \bs{a} + (\bs{\beta}^\star)^\tr \bs{b}$,
we have
\begin{equation}
\OTphi(\va, \vb) \le \OT(\bs{a}, \bs{b})
- \frac{\gamma}{2} ( \|\valpha_\Phi\|^2 + \|\vbeta_\Phi\|^2 ).
\end{equation}
Likewise,
\begin{equation}
\OT(\bs{a}, \bs{b}) - 
\frac{\gamma}{2} (\|\valpha^\star\|^2 + \|\vbeta^\star\|^2) 
\le \OTphi(\bs{a}, \bs{b}).
\end{equation}
Combining the two, we get
\begin{equation}
\OT(\bs{a}, \bs{b}) - 
\frac{\gamma}{2} (\|\valpha^\star\|^2 + \|\vbeta^\star\|^2) 
\le
\OTphi(\va, \vb) \le \OT(\bs{a}, \bs{b})
- \frac{\gamma}{2} ( \|\valpha_\Phi\|^2 + \|\vbeta_\Phi\|^2 ).
\label{eq:bound_ot_phi_l2}
\end{equation}
Hence we need to bound variables $\valpha,\vbeta \in \mathcal{P}(C)$.  Since
$\normtwo{\cdot} \le \normone{\cdot}$, we can upper bound
$\normone{\valpha^\star} + \normone{\vbeta^\star}$. In addition, we can always
add the additional constraint that $\valpha^\tr \va + \beta^\tr \vb \ge
\mathbf{0}^\tr \va + \mathbf{0}^\tr \vb = 0$ since $(\mathbf{0}, \mathbf{0})$ is
dual feasible for \eqref{eq:dual_lp}.  Since for any optimal pair
$\valpha^\star,\vbeta^\star$, the pair
$\valpha^\star-\sigma\mathbf{1},~\vbeta^\star+\sigma\mathbf{1}$ is also feasible
and optimal for any $\sigma \in \mathbb{R}$, we can also add
the constraint $\valpha^\tr \bs{1}_m = 0$. The obtained bound will obviously hold for
any optimal pair $\valpha^\star,\vbeta^\star$. Hence, we can apply Lemma
\ref{lemma:bound_1norm_PC}.  By the same reasoning but using the constraint
$\vbeta^\tr \bs{1}_n = 0$ in place of $\valpha^\tr \bs{1}_m = 0$, we can obtain
a similar bound. By combining these two bounds, we obtain our final bound:
\begin{equation}
\|\valpha\|_1 + \|\vbeta\|_1 \le \|C\|_\infty \min\{\nu_1 + n,
\nu_2 + m\}
\end{equation}
where
\begin{equation}
\begin{split}
\nu_1 & = \max \left\{ (2+n/m)\norminf{\va^{-1}}, \norminf{\vb^{-1}}\right\} \\
    \nu_2 & = \max \left\{\norminf{\va^{-1}},
    (2+m/n)\norminf{\vb^{-1}}\right\}.
\end{split}
\end{equation}
Taking the square of this bound and plugging the result in
\eqref{eq:bound_ot_phi_l2} gives the claimed result. 
Applying the same reasoning with Lemma \ref{lemma:bound_1norm_PC_semi} gives the
claimed result for the semi-relaxed primal.

\section{Alternating minimization with exact block updates}
\label{appendix:alt_min}

\textbf{General case.} Let $\bs{\beta}(\bs{\alpha})$ be an optimal
solution of \eqref{eq:smoothed_dual} given $\bs{\alpha}$ fixed, and similarly
for $\bs{\alpha}(\bs{\beta})$.  From the first-order optimality conditions,
\begin{align}
    \nabla \indomega\left(\bs{\alpha} + \beta_j(\bs{\alpha}) \bs{1}_m - \col{c}{j}
\right)^\tr \bs{1}_m &= b_j \quad \forall j \in [n]
\label{eq:opt_condition_beta}
\end{align}
and similarly for $\bs{\alpha}$ given $\bs{\beta}$ fixed. Solving these
equations is non-trivial in general. However, because
\begin{equation}
    \nabla \indomega\left(\bs{\alpha} + \beta_j(\bs{\alpha}) \bs{1}_m - \col{c}{j}
    \right) = b_j \nabla \maxopj(\bs{\alpha} - \col{c}{j})
\end{equation}
holds $\forall \bs{\alpha} \in \mathbb{R}^m$, $j \in [n]$, we can retrieve
$\beta_j(\bs{\alpha})$ if we know how to compute 
$\nabla \maxop(\bs{x})$ and
the inverse map $(\nabla \indomega)^{-1}(\bs{y})$ exists.
That map exists and equals $\nabla \Omega(\bs{y})$ provided that
$\Omega$ is differentiable and $\bs{y} > \bs{0}$.  

\textbf{Entropic regularization.}
It is easy to verify that \eqref{eq:opt_condition_beta} is satisfied with
\begin{equation}
\bs{\beta}(\bs{\alpha}) = \gamma \log\left(
\frac{\bs{b}}{K^\tr \expdivgamma{\bs{\alpha}}{-\bs{1}_m}}\right)
\quad \text{where} \quad
K \coloneqq \expdivgamma{-C}{}
\end{equation}
and similarly for $\bs{\alpha}(\bs{\beta})$. 
These updates recover the iterates of the Sinkhorn algorithm
\citep{sinkhorn_distances}.

\textbf{Squared $2$-norm regularization.}
Plugging the expression of $\nabla \indomega$ in \eqref{eq:opt_condition_beta},
we get that $\bs{\beta}(\bs{\alpha})$ must satisfy
\begin{equation}
[\bs{\alpha} + \beta_j(\bs{\alpha}) \bs{1}_m - \col{c}{j}]_+^\tr 
\bs{1}_m = \gamma b_j \quad \forall j \in [n].
\end{equation}
Close inspection shows that it is exactly the same optimality condition as the Euclidean projection onto
the simplex $\displaystyle{\argmin_{\bs{y} \in \Simplex^m}} \|\bs{y} - \bs{x}\|^2$
must satisfy, with $\bs{x} = \frac{\bs{\alpha} - \col{c}{j}}{\gamma b_j}$.  Let
$x_{[1]} \ge \dots \ge x_{[m]}$ be the values of $\bs{x}$ in sorted order.
Following \citep{michelot,duchi}, if we let
\begin{equation}
\rho \coloneqq \max \left\{i \in [m] \colon x_{[i]} - 
\frac{1}{i} \left( \sum_{r=1}^{i} x_{[r]} - 1 \right) > 0 \right\}
\end{equation}
then $\bs{y}^\star$ is \textit{exactly} achieved at $[\bs{x} +
\frac{\beta_j(\bs{\alpha})}{\gamma b_j} \bs{1}_m]_+$, where
\begin{equation}
\beta_j(\bs{\alpha}) = -\frac{\gamma b_j}{\rho} \left( \sum_{r=1}^{\rho} x_{[r]}
- 1 \right).
\end{equation}
The expression for $\bs{\alpha}(\bs{\beta})$ is completely symmetrical.  
While a projection onto the simplex is required
for each coordinate, as discussed in \S\ref{sec:closed_form_expr},
this can be done in expected linear time.
In addition, each coordinate-wise solution can be computed in parallel.

\textbf{Alternating minimization.} Once we know how to compute
$\bs{\beta}(\bs{\alpha})$ and $\bs{\alpha}(\bs{\beta})$, there are a number of
ways we can build a proper algorithm to solve the smoothed dual. Perhaps the
simplest is to alternate between $\bs{\beta} \leftarrow \bs{\beta}(\bs{\alpha})$
and $\bs{\alpha} \leftarrow \bs{\alpha}(\bs{\beta})$.  For entropic
regularization, this two-block coordinate descent (CD) scheme is known as the
Sinkhorn algorithm and was recently popularized in the context of optimal
transport by \citet{sinkhorn_distances}. A disadvantage of this approach,
however, is that computational effort is spent updating coordinates that may
already be near-optimal. To address this issue, we can instead adopt a greedy CD
scheme as recently proposed for entropic regularization by \citet{greenkhorn}.

\section{Additional experiments}
\label{appendix:exp}

We ran the same experiments as Figure \ref{fig:color_transfer_cmp} and Figure
\ref{fig:convergence} on one more image pair:
``Grafiti'' by Jon Ander and ``Rainbow Bridge National Monument Utah'', by
Bernard Spragg. Both images are in the public domain.  The results, presented in
Figure \ref{fig:convergence2} and Figure \ref{fig:color_transfer_cmp2} below,
confirm the empirical findings described in \S\ref{sec:app_color_transfer} and
\S\ref{sec:solver_cmp}.
The images are available at
\url{https://github.com/mblondel/smooth-ot/tree/master/data}.
\begin{figure*}[p]
\centering
\includegraphics[scale=0.38]{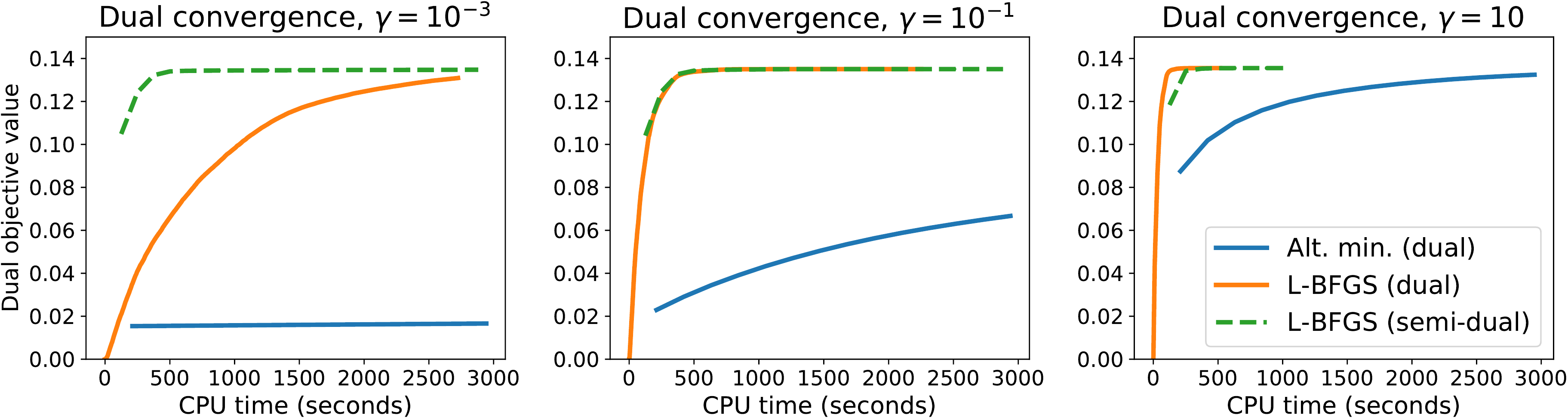} \\[1em]
\caption{Same experiment as Figure \ref{fig:convergence} on one more image
pair.}
\label{fig:convergence2}
\end{figure*}

\begin{figure*}[p]
\centering
\includegraphics[scale=0.72]{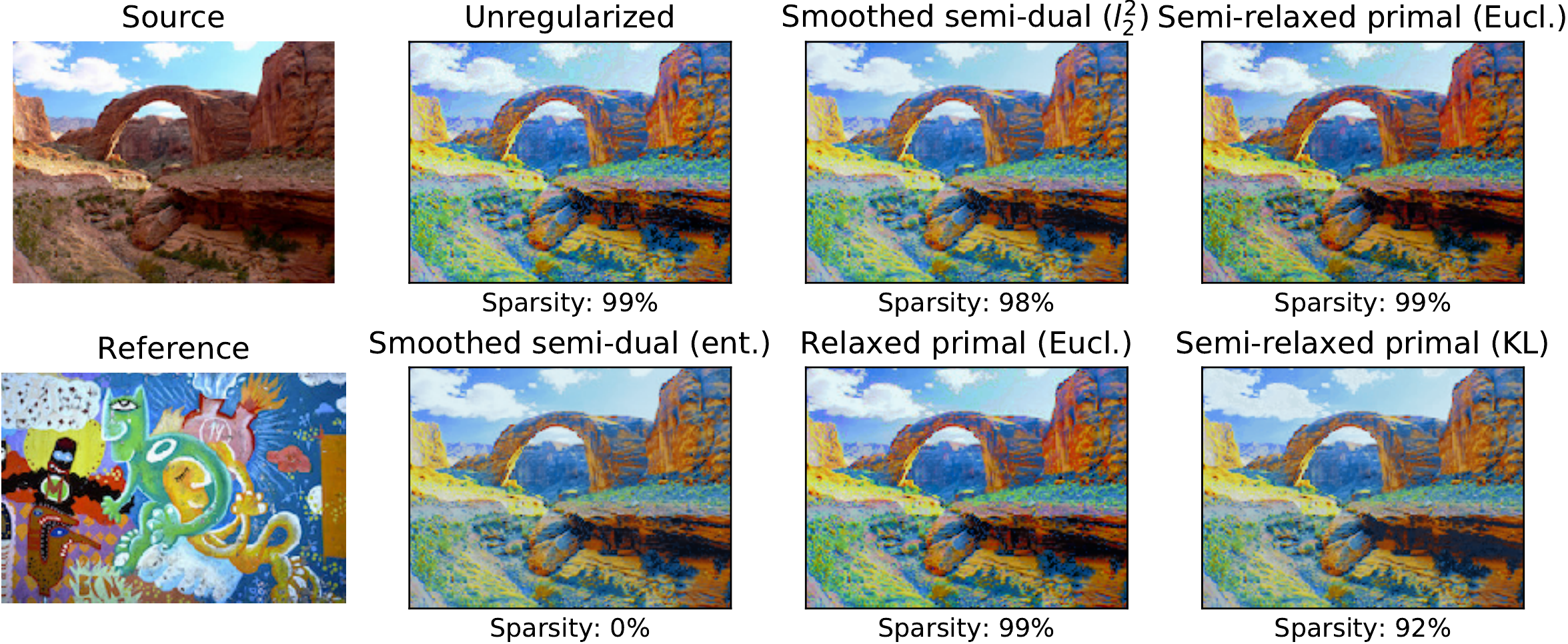} \\[3em]
\caption{Same experiment as Figure \ref{fig:color_transfer_cmp} on one more
image pair.}
\label{fig:color_transfer_cmp2}
\end{figure*}

\end{document}